\documentclass{article}

\usepackage{microtype}
\usepackage{graphicx}
\usepackage{subcaption}
\usepackage{booktabs} 
\usepackage{multirow}
\usepackage{bm}
\usepackage[normalem]{ulem}
\useunder{\uline}{\ul}{}

\usepackage{setspace}
\usepackage{threeparttable}
\usepackage{url}
\usepackage{makecell}

\usepackage{hyperref}

\usepackage[arxiv]{icml2024}

\usepackage{amsmath}
\usepackage{amssymb}
\usepackage{mathtools}
\usepackage{amsthm}

\usepackage[capitalize,noabbrev]{cleveref}

\usepackage{amsmath,amsfonts,bm}

\def\eqref#1{equation~\ref{#1}}

\def\1{\bm{1}}

\def\eps{{\epsilon}}

\def\vc{{\bm{c}}}

\def\vf{{\bm{f}}}
\def\vg{{\bm{g}}}

\def\vv{{\bm{v}}}

\def\vx{{\bm{x}}}

\def\vz{{\bm{z}}}

\DeclareMathAlphabet{\mathsfit}{\encodingdefault}{\sfdefault}{m}{sl}
\SetMathAlphabet{\mathsfit}{bold}{\encodingdefault}{\sfdefault}{bx}{n}

\theoremstyle{plain}

\theoremstyle{definition}

\theoremstyle{remark}

\usepackage[textsize=tiny]{todonotes}

\icmltitlerunning{Trajectory Consistency Distillation}

\begin{document}

\twocolumn[
\icmltitle{\textsc{Trajectory Consistency Distillation}: Improved Latent Consistency Distillation by Semi-Linear Consistency Function with Trajectory Mapping}



\icmlsetsymbol{equal}{*}

\begin{icmlauthorlist}
\icmlauthor{Jianbin Zheng}{equal,scut}
\icmlauthor{Minghui Hu}{equal,ntu}
\icmlauthor{Zhongyi Fan}{bit}
\icmlauthor{Chaoyue Wang}{usyd} \\
\icmlauthor{Changxing Ding}{scut} 
\icmlauthor{Dacheng Tao}{ntu}
\icmlauthor{Tat-Jen Cham}{ntu}
\end{icmlauthorlist}

\icmlaffiliation{scut}{South China University of Technology}
\icmlaffiliation{ntu}{Nanyang Technological University}
\icmlaffiliation{usyd}{The University of Sydney}
\icmlaffiliation{bit}{Beijing Institute of Technology}

\icmlcorrespondingauthor{Minghui Hu}{e200008@e.ntu.edu.sg}

\icmlkeywords{Diffusion, Consistency Models, Text-to-Image Generation}

\vskip 0.15in

\begin{center}
    Project Page: \url{https://mhh0318.github.io/tcd}
\end{center}
\vskip -0.15in
{
\begin{figure}[H]
    \centering
    \includegraphics[width=\textwidth]{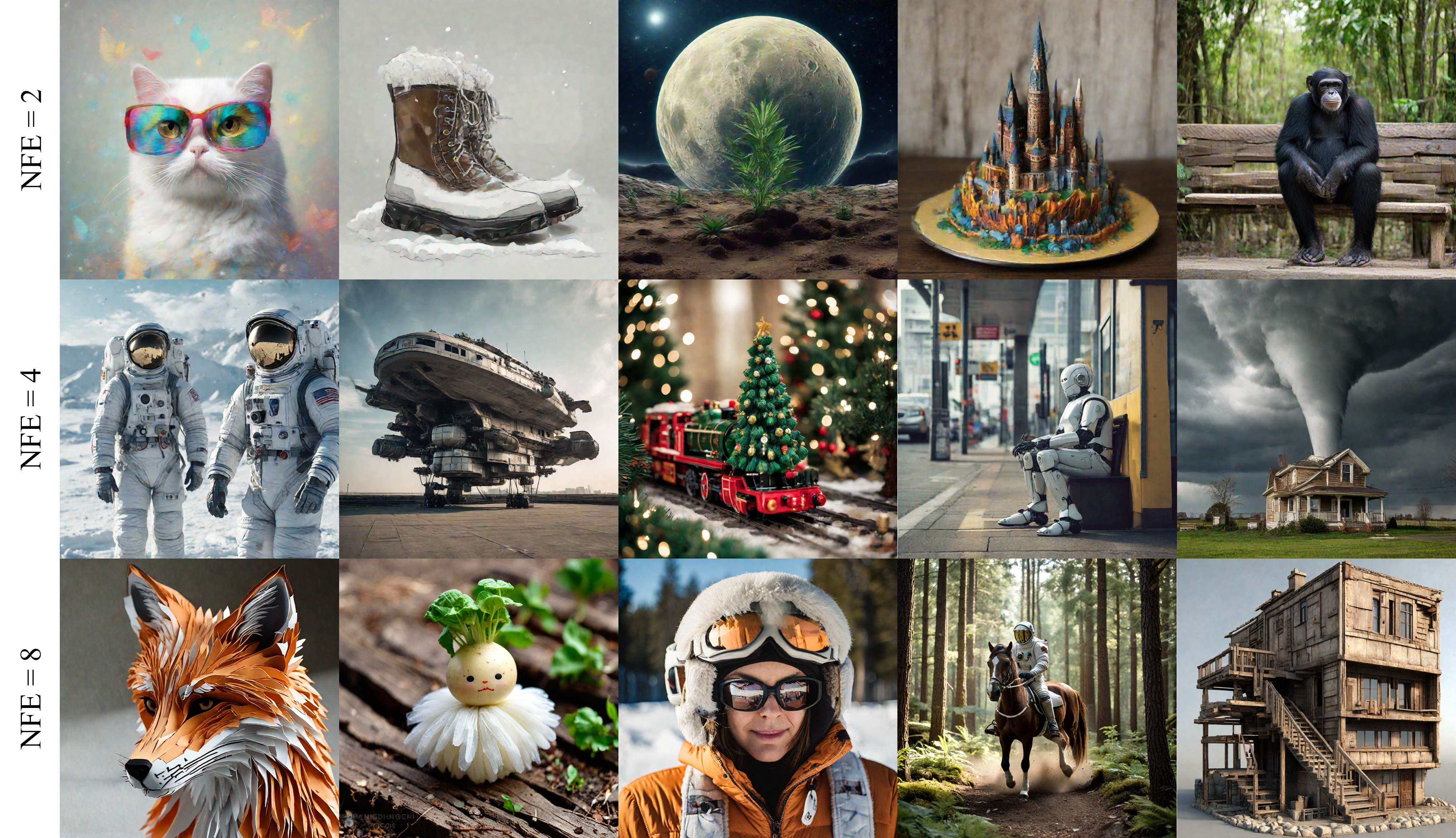}
\end{figure}

\vskip 0.1in
}

]



\printAffiliationsAndNotice{\icmlEqualContribution} 

\begin{abstract}
Latent Consistency Model (LCM) extends the Consistency Model to the latent space and leverages the guided consistency distillation technique to achieve impressive performance in accelerating text-to-image synthesis.
However, we observed that LCM struggles to generate images with both clarity and detailed intricacy.
Consequently, we introduce Trajectory Consistency Distillation (TCD), which encompasses \emph{trajectory consistency function} and \emph{strategic stochastic sampling}.
The trajectory consistency function diminishes the parameterisation and distillation errors by broadening the scope of the self-consistency boundary condition with trajectory mapping and endowing the TCD with the ability to accurately trace the entire trajectory of the Probability Flow ODE in semi-linear form with an Exponential Integrator.
Additionally, strategic stochastic sampling provides explicit control of stochastic and circumvents the accumulated errors inherent in multi-step consistency sampling.
Experiments demonstrate that TCD not only significantly enhances image quality at low NFEs but also yields more detailed results compared to the teacher model at high NFEs.

\end{abstract}

\section{Introduction}

Score-based generative models (SGMs), also commonly known as Diffusion Models~\cite{sohl2015deep, song2019generative, song2020score, ho2020denoising}, have demonstrated their proficiency in various generative modeling domains such as image~\cite{dhariwal2021diffusion, ramesh2022hierarchical, rombach2022high}, video~\cite{ho2020denoising, wu2023tune, guo2023animatediff}, and audio~\cite{kong2020diffwave, chen2020wavegrad, popov2021grad}, particularly in text-to-image synthesis~\cite{nichol2022glide, ramesh2022hierarchical, saharia2022photorealistic, podell2023sdxl}. An noteworthy aspect of SGMs is the utilisation of stochastic differential equations (SDEs) and corresponding marginal-preserving ordinary differential equations (ODEs) to iteratively perturb data and eliminate noise~\cite{song2020score}. This facilitates an effective trade-off between generation cost and sampling quality, but they are also constrained by slow inference speed, requiring a substantial number of function evaluations (NFEs) to obtain satisfactory results.

To overcome this limitation, \citet{song2023consistency} proposed Consistency Models (CMs), an emerging class of powerful generative models capable of generating high-quality data with single-step or few-step sampling without the need for adversarial training. CMs are inextricably connected with SGMs in their underlying mathematical underpinnings, aiming to enforce the self-consistency property by mapping arbitrary points on the trajectory of the same Probability Flow Ordinary Differential Equation (PF ODE) to the trajectory's origin~\cite{song2020score}. CMs can be trained with consistency distillation or treated as standalone generative models. \citet{song2023consistency} have demonstrated their superiority through extensive experiments in the pixel space. Latent Consistency Models (LCMs)~\cite{luo2023latent} further incorporate Latent Diffusion Models (LDMs)~\cite{rombach2022high} with CMs, achieving remarkable success in swiftly synthesizing high-resolution images conditioned on text. Moreover, LCM-LoRA~\cite{luo2023lcm} improves the training efficiency of LCMs and converts them into a universal neural PF ODE solver by introducing LoRA~\cite{hu2021lora} into the distillation process of LCMs. It is noteworthy that all these Consistency-Type Models still allow for striking a balance between computation and sample quality using Multistep Consistency Sampling~\cite{song2023consistency}. There are also some works that focus on changing the interval of the consistency function or the distillation learning protocol~\cite{gu2023boot, berthelot2023tract,heek2024multistep}, leading to enhanced outcomes.

\begin{figure}[ht!]
    \centering
    \includegraphics[width=\linewidth]{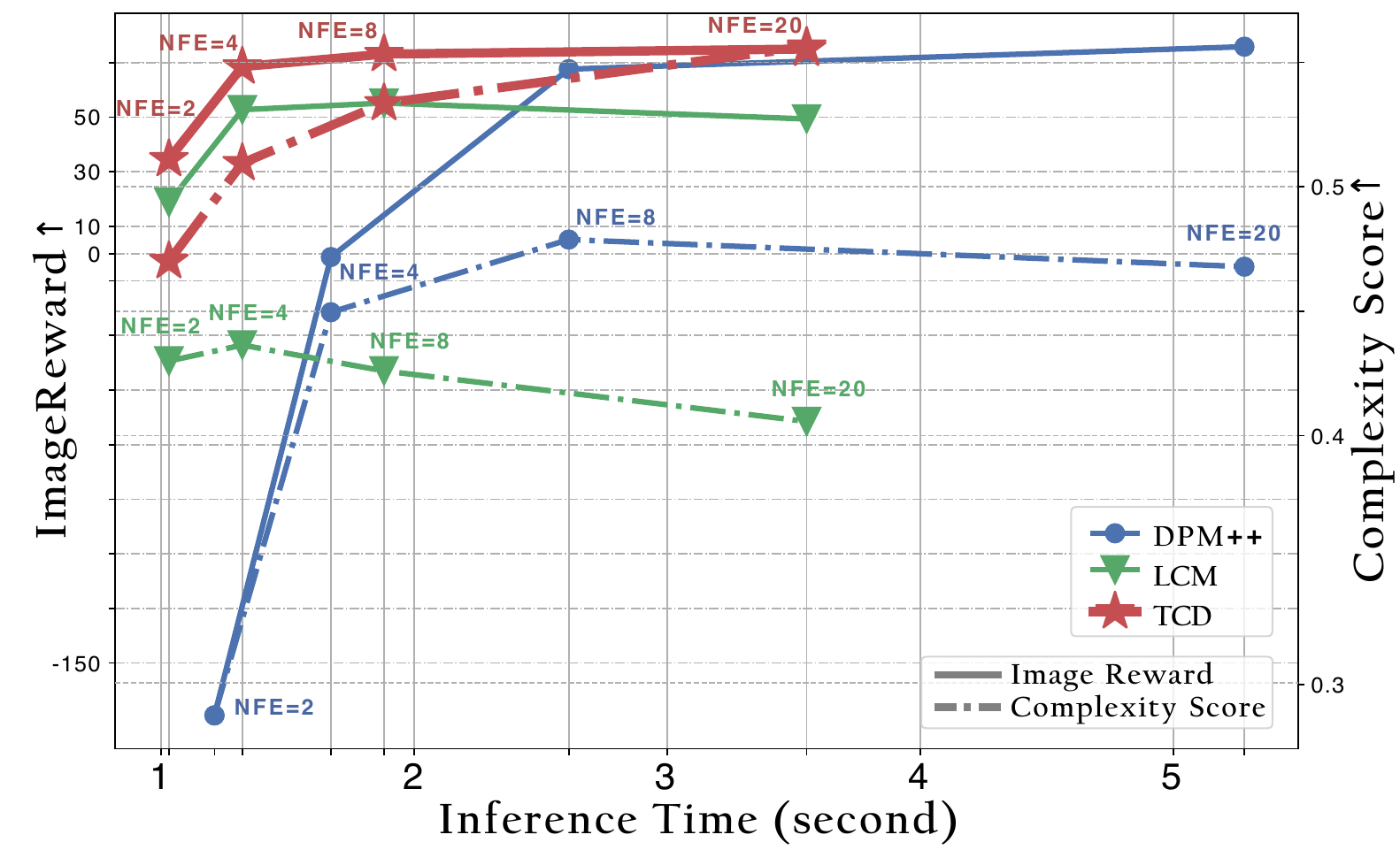}
    \caption{Comparison between TCD and other state-of-the-art methods. TCD delivers exceptional results in terms of both quality and speed, completely surpassing LCM. Notably, LCM experiences a notable decline in quality at high NFEs. In contrast, TCD maintains superior generative quality at high NFEs, even exceeding the performance of DPM-Solver++(2S) with origin SDXL.}
    \label{fig:teaser}
\end{figure}

Despite the introduction of the guided distillation method and skipping-step technique~\cite{luo2023latent} by LCMs for effectively distilling knowledge from pre-trained diffusion models, the quality of images generated by LCMs in a single step or even with minimal steps (4$\sim$8) still lags significantly behind the convergence of its teacher model. 
Our investigation revealed that, in practice, increasing the number of inference iterations diminishes the visual complexity and quality of the results, as illustrated in~\cref{fig:teaser}.
This renders LCMs less capable of synthesizing samples that are perceptually comparable to those of their teacher model.
Recent findings by EDM~\cite{karras2022elucidating} identified that an excessive Langevin-like addition and removal of random noise result in a gradual loss of detail in the generated images. Additionally,~\citet{li2023diffusion} showed evidence that during the process of multi-step sampling, discretisation errors accumulate over iterations, ultimately causing the generated images to deviate from the target distribution. Recently, in order to balance the inference steps and the sample quality during inference, CTM~\cite{kim2023consistency} proposed a unified framework that takes advantage of the sampling methods of Distillation Models~\cite{salimans2022progressive, song2023consistency} and SGMs according to a novel representation of denoisers and analyzed the errors during sampling.

In this paper, we focus on improving the generation quality of LCMs, especially reducing the errors arising during parameterisation process of LCMs. We introduce Trajectory Consistency Distillation (TCD), as summarized in~\cref{fig:pipeline}, which comprises two key elements: \emph{trajectory consistency function} and \emph{strategic stochastic sampling}. Specifically, inspired by the form of exponential integrators~\cite{lu2022dpm, lu2022dpm++}, the trajectory consistency function (TCF) diminishes parameterisation and distillation errors by expanding the boundary conditions of the consistency model and enabling seamless transitions at any point along the trajectory governed by the PF ODE, as demonstrated in~\cref{fig:tcf}. Furthermore, strategic stochastic sampling (SSS) allows control of stochasticity and suppresses the accumulated discretisation error and estimation error according to the narrowed bi-directional iterations as shown in~\cref{fig:sss}.

Experiments show that TCD can significantly enhance the quality of images generated by LCM, surpassing it in performance. Furthermore, TCD is capable of outperforming the teacher model (\textit{e.g.} SDXL with DPMSolver++) when sampling with sufficient iterations (\textit{e.g.}, 20 NFEs).

\section{Related Works}
\paragraph{Diffusion Models.}
Diffusion Models (DMs), also known as Score-based generative models, have shown superior performance in various generative fields. \citet{chen2022sampling} provides theoretical convergence guarantees, implying that DMs can efficiently sample from essentially any realistic data distribution under minimal data assumptions.
ADM~\cite{dhariwal2021diffusion} firstly shows the potential of DMs to outperform GANs. EDM~\cite{karras2022elucidating} further elucidates the design space of DMs, clearly separating the concrete design choices, deriving best practices for the sampling process, and improving the training dynamics, thereby drastically improving the results.
\paragraph{Text-conditional Diffusion Models.}
DMs have specially achieved great success in the area of text-to-image synthesis~\cite{nichol2022glide, ramesh2022hierarchical, rombach2022high, saharia2022photorealistic, balaji2022ediffi, podell2023sdxl}.
To reduce computational cost, diffusion models typically operate within the latent space~\cite{rombach2022high, podell2023sdxl} or include separate super-resolution steps~\cite{ramesh2022hierarchical, saharia2022photorealistic, balaji2022ediffi}.
The integration of classifier-free diffusion guidance~\cite{ho2022classifier, dieleman2022guidance} during the sampling process dramatically improves samples produced by conditional diffusion models at almost no additional cost.
\paragraph{Fast Sampling of DMs.}
DMs exhibit great generative capabilities but are bottlenecked by their slow sampling speed. Various numerical methods driven by the SDE mathematical underpinnings underlying DMs have been proposed for accelerating DM sampling.
DDIM~\cite{song2020denoising} has originally shown promise for few-step sampling. Other works involve predictor-corrector samplers~\cite{song2020score, karras2022elucidating}, exponential integrators~\cite{lu2022dpm, zhang2022fast, lu2022dpm++}, and automated methods to adjust solvers~\cite{kong2021fast, jolicoeur2021gotta}. Among them, the semi-linear structure utilizing exponential integrators~\cite{lu2022dpm, zhang2022fast, lu2022dpm++} can derive accurate solutions for the PF-ODEs. By separating the linear and nonlinear parts, it skillfully bypasses the inherent discretization errors in the linear items, thereby obtaining exact analytical solutions for the linear term.
Another series of research works exemplified by distillation techniques~\cite{luhman2021knowledge, salimans2022progressive, meng2023distillation}. These methods distill knowledge from pretrained models into few-step samplers, representing an efficient solution for few NFEs. However, they may experience a lengthy and costly process that requires huge amounts of data and suffer from slow convergence.
\paragraph{Consistency Models.} 
To overcome the limitations of current fast samplers, \citet{song2023consistency} proposes Consistency Models (CMs) learning a direct mapping from noise to data built on the top of the trajectory of PF ODE, achieving one-step generation while allowing multi-step sampling to trade compute for quality. \citet{lyu2023convergence} provide the first convergence guarantees for CMs under moderate assumptions. 
\citet{kim2023consistency} proposes a universal framework for CMs and DMs samplings with an ``anytime to anytime" decoder, which is able to predict both infinitesimally small steps and arbitrary long step jumps. Together with a $\gamma$-sampler, it achieves a balance between sampling quality and computational resources. Both CTM and TCD eliminate constraints on predictions at the origin during distillation, but from distinct viewpoints, leading to different solutions. Furthermore, CTM provides $\gamma$ sampling with a well-founded theoretical basis in the view of multistep sampling, while our experiments disseminate its theoretical findings.
Further comments about CTM and TCD can be found in Appendix~\ref{ctmd}.
Latent Consistency Models (LCMs)~\cite{luo2023latent} integrate consistency distillation with latent diffusion models~\cite{rombach2022high} to achieve impressive performance in accelerating text-to-image synthesis. LCM-LoRA~\cite{luo2023lcm} further improves the training efficiency and versatility of LCMs by introducing LoRA~\cite{hu2021lora} into the distillation process. Several studies have concentrated on modifying the interval of the consistency function and refining the distillation learning protocol~\cite{gu2023boot, berthelot2023tract, heek2024multistep}, resulting in improved results.
\paragraph{Parameter-Efficient Fine-Tuning.}
Training a diffusion model is highly resource-intensive and environmentally unfriendly. Fine-tuning such a model can also be challenging due to the vast number of parameters involved~\cite{aghajanyan2020intrinsic}.
Thus, Parameter-Efficient Fine-Tuning (PEFT)~\cite{houlsby2019parameter} was proposed to enable the fine-tuning of pretrained models with a limited number of parameters required for training.
Among these techniques, Low-Rank Adaptation (LoRA)~\cite{hu2021lora} has demonstrated transcendental performance. LoRA's strategy involves freezing the pretrained model weights and injecting trainable rank decomposition matrices, which succinctly represent the required adjustments in the model's weights for fine-tuning. With this strategy, LoRA significantly reduces the volume of parameters to be modified, thereby substantially decreasing both computational load and storage demands.

\section{Preliminaries}
\subsection{Diffusion Models}
Diffusion Models (DMs) start with a predefined forward process $\{\bm{x}_t\}_{t\in[0, T]}$ indexed by a continuous time variable $t$ with $T > 0$, which progressively adds noise to data via Gaussian perturbations. The forward process can be modeled as a widely used stochastic differential equation (SDE)~\cite{song2020score, karras2022elucidating}:
\begin{equation} \label{eq:sde}
    \text{d}\bm{x}_t = \mu(t)\bm{x}_t\text{d}t + \nu(t)\text{d}\bm{w}_t,
\end{equation}
where $\bm{w}_t$ denotes the $d$-dimensional standard Brownian motion and $\mu(t)\colon\mathbb{R}\rightarrow\mathbb{R}$ and $\nu(t)\colon\mathbb{R}\rightarrow\mathbb{R}$ are the drift and diffusion coefficients, respectively, where $d$ is the dimensionality of the dataset.
Denote the marginal distribution of $\bm{x}_t$ following the forward process as $p_t(\bm{x}_t)$ and, such an It\^o SDE gradually perturbs the empirical data distribution $p_0(\bm{x}) = p_{\rm{data}}(\bm{x})$ towards the prior distribution $p_T(\bm{x}) \approx \pi(\bm{x})$ approximately, where $\pi(\bm{x})$ is a tractable Gaussian distribution.

Remarkably, \citet{song2020score} proved that there exists an ordinary differential equation (ODE) dubbed the \textit{probability flow} (PF) ODE, whose trajectories share the same marginal probability densities $\{p_t(\bm{x})\}_{t\in[0, T]}$ as the forward SDE,
\begin{equation} \label{eq:pfode}
    \frac{\text{d}\bm{x}_t}{\text{d}t} = \mu(t)\bm{x}_t - \frac{1}{2} \nu(t)^2 \nabla_{\bm{x}} \log p_t(\bm{x}_t).
\end{equation}
As for sampling, the ground truth score in~\cref{eq:pfode} is approximated with the learned score model $\bm{s}_{\bm{\theta}}(\bm{x}, t) \approx \nabla_{\bm{x}} \log p_t(\bm{x})$ via score matching~\cite{hyvarinen2009estimation, song2019generative, ho2020denoising}. This yields an empirical estimate of the PF ODE, referred to as the \textit{empirical PF ODE}:
\begin{equation} \label{eq:emp_pfode}
    \frac{\text{d}\tilde{\bm{x}}_t}{\text{d}t} = \mu(t)\tilde{\bm{x}}_t - \frac{1}{2} \nu(t)^2 \bm{s}_{\bm{\theta}}(\tilde{\bm{x}}_t, t).
\end{equation}
Then samples can be drawn by solving the empirical PF ODE from $T$ to 0. There already exist off-the-shelf ODE solvers~\cite{song2020score, song2020denoising, karras2022elucidating} or efficient numerical solvers~\cite{lu2022dpm, lu2022dpm++, zhang2022fast} that can be directly applied to approximate the exact solution.

\begin{figure}[t!]
    \centering
    \begin{subfigure}{\linewidth}
        \includegraphics[width=\linewidth]{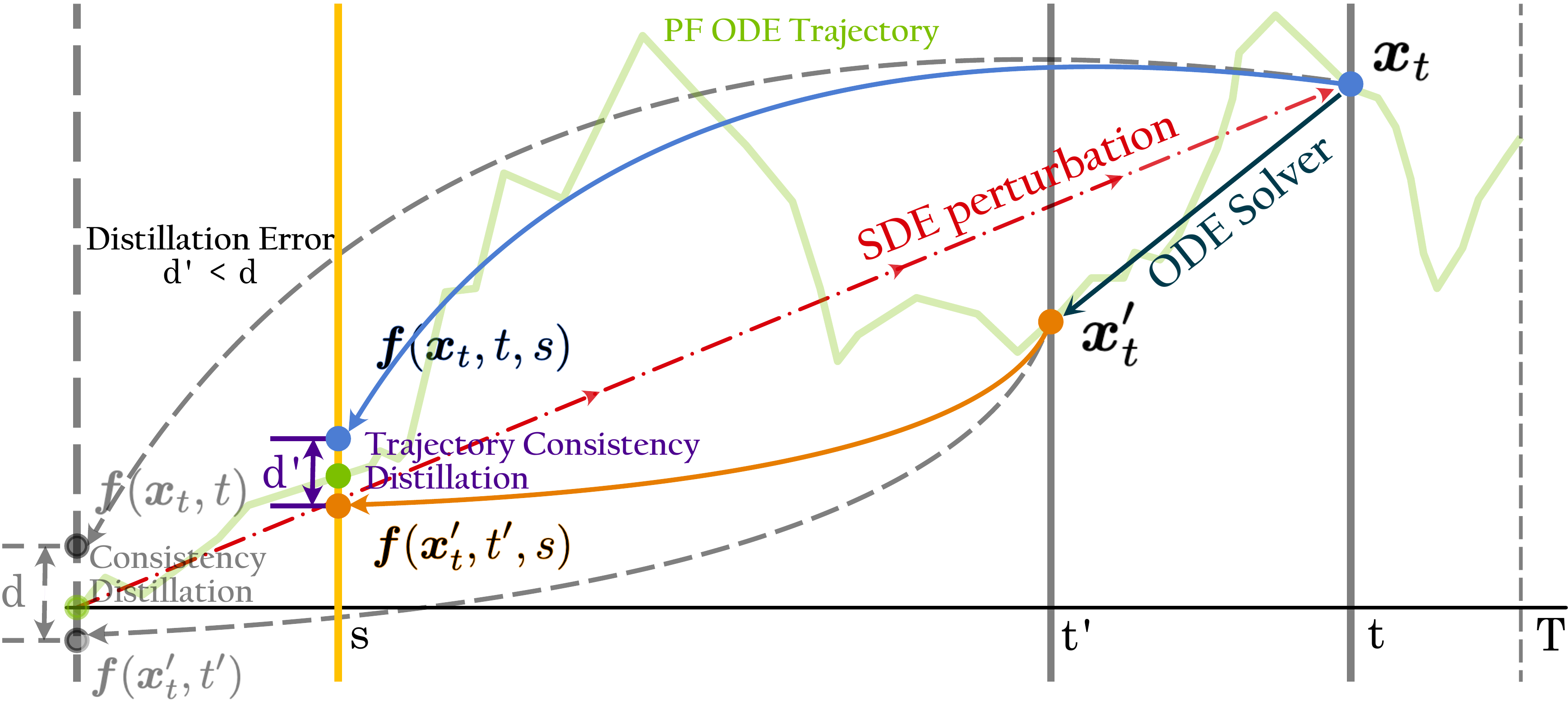}
        \caption{Training process, wherein the TCF expands the boundary conditions to an arbitrary timestep of $s$, thereby reducing the theoretical upper limit of error. The details can be found in~\cref{alg:tcd}.}
        \label{fig:tcf}
    \end{subfigure}
    \begin{subfigure}{\linewidth}
        \includegraphics[width=\linewidth]{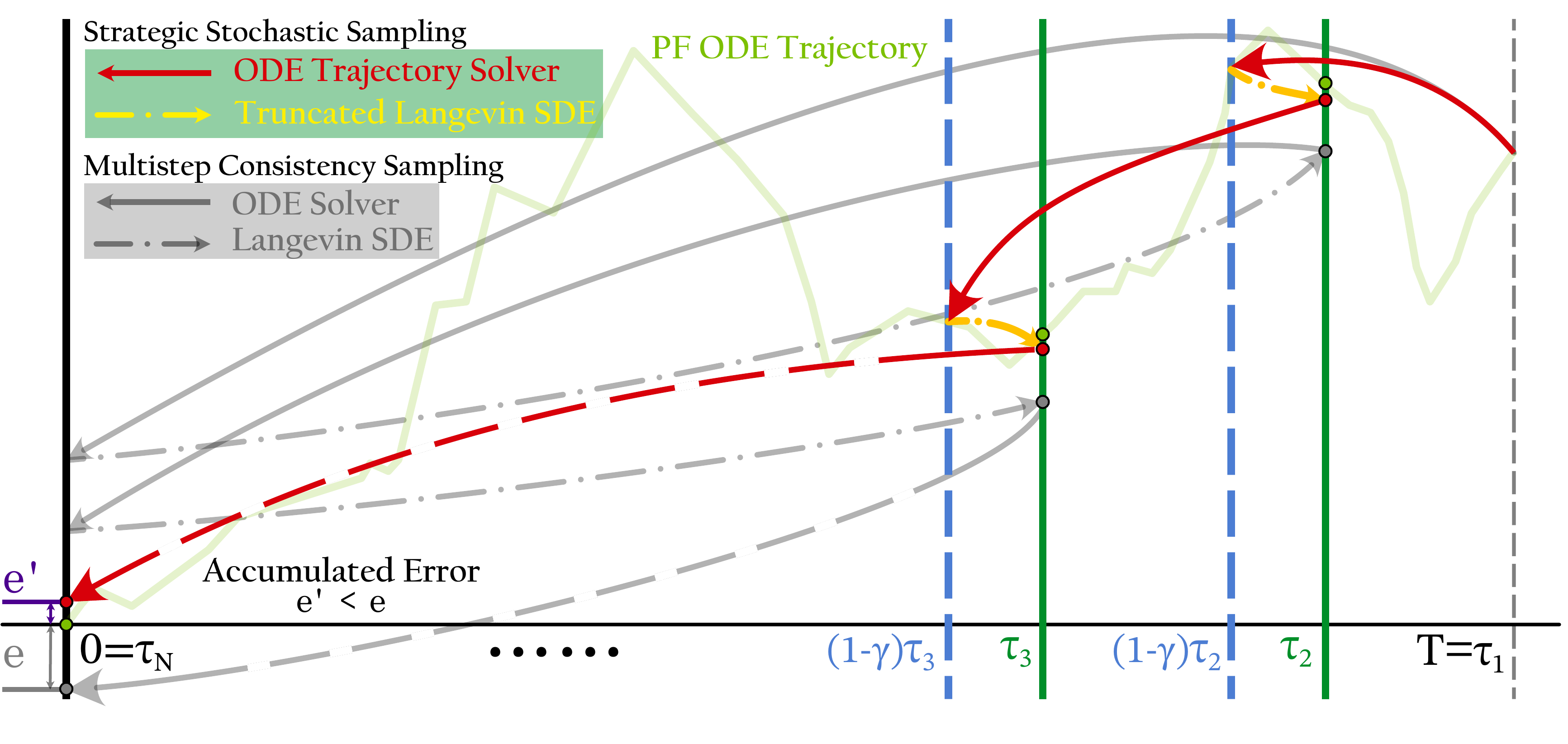}
        \caption{Sampling process, as detailed in~\cref{alg:mcs}~\&~\cref{alg:sss}.}
        \label{fig:sss}
    \end{subfigure}
    \caption{The comparative overview of the baseline Consistency Distillation~\cite{song2023consistency} and the proposed Trajectory Consistency Distillation, includes Trajectory Consistency Function (TCF) for training and Strategic Stochastic Sampling (SSS) for inference.}
    \label{fig:pipeline}
\end{figure}

\subsection{Consistency Models}
Solving~\cref{eq:emp_pfode} typically involves numerous neural network evaluations to generate decent samples. Thus, consistency models are proposed to directly map any points along the trajectory $\{\bm{x}_t\}_{t\in[0, T]}$ of the PF ODE to the origin of its trajectory, thereby facilitating generation in just a few steps. The associated mapping can be formuled as follows:
\begin{equation} \label{eq:consistency_f}
    \bm{f}(\bm{x}_t, t) = \bm{x}_0 \quad \forall  t\in[0, T],
\end{equation}
with the boundary condition $\bm{f}(\bm{x}_0, 0) = \bm{x}_0$. It is worth noting that~\cref{eq:consistency_f} is equivalent to the \textit{self-consistency} condition:
\begin{equation} \label{eq:consistency_prop}
    \bm{f}(\bm{x}_t,t)=\bm{f}(\bm{x}_{t'},t')\quad \forall  t, t'\in[0, T].
\end{equation}

A parametric model $\bm{f}_{\bm{\theta}}$ is constructed to estimate the consistency function $\bm{f}$ by enforcing the self-consistency property.
Typically, $\bm{f}$ can be distilled from a pretrained diffusion model $F_{\bm{\theta}}(\bm{x}_t, t)$ and parameterized as:
\begin{equation} \label{eq:generalform}
    \bm{f}_{\bm{\theta}}(\bm{x}_t, t) =
\begin{cases}
\bm{x}_0, & t=0 \\
F_{\bm{\theta}}(\bm{x}_t, t), & t\in(0, T]
\end{cases}
\end{equation}
or using skip-connection as:
\begin{equation} \label{eq:cf_cm}
    \vf_{\bm{\theta}} (\vz_t, \vc,  t) = c_{\text{skip}}(t) \vz_t + c_{\text{out}}(t) F_{\bm{\theta}}(\bm{x}_t, t),
\end{equation}
where  $c_{\text{skip}}(\epsilon)=1$ and $c_{\text{out}}(\epsilon)=0$.

For training consistency models, the objective of Consistency Distillation (CD) is defined as minimizing:
\begin{equation}
\begin{split}
    \mathcal{L}_{\rm{CD}}^{N}&(\bm{\theta}, \bm{\theta^-}; \bm{\phi}) := \\
    & \mathbb{E} \left[\lambda(t_n) \left\lVert{\bm{f}_{\bm{\theta}}(\bm{x}_{t_{n+1}}, t_{n+1}) - \bm{f}_{\bm{{\theta^-}}}(\hat{\bm{x}}_{t_n}^{\bm{\phi}}, t_n)}\right\rVert_2^2 \right], 
\end{split}
\end{equation}
where $0 = t_1 < t_2 \cdots < t_N = T$, $n$ uniformly distributed over $\{1, 2, \cdots , N-1\}$, $\lambda(\cdot) \in \mathbb{R}^+$ is a positive weighting function, and the expectation is taken with respect to $\bm{x} \sim p_{\rm{data}}$. $\bm{x}_{t_{n+1}}$ can be sampled using SDE~(\ref{eq:sde}) and $\hat{\bm{x}}_{t_n}^{\bm{\phi}}$ is calculated by $\hat{\bm{x}}_{t_n}^{\bm{\phi}} := \Phi(\bm{x}_{t_{n+1}}, t_{n+1}, t_n; \bm{\phi})$, where $\Phi(\cdots; \bm{\phi})$ represents the update function of a one-step ODE solver applied to the empirical PF ODE~(\ref{eq:emp_pfode}). Moreover, $\bm{\theta^-}$ is introduced to stabilize the training process and updated by an exponential moving average (EMA) strategy, \textit{i.e.}, given $0\leq \mu<1$, $\bm{\theta}^- \leftarrow \texttt{sg}( \mu\bm{\theta}^-+(1-\mu)\bm{\theta})$.
With a well-trained model $\bm{\theta^*}$ such that $\mathcal{L}_{\rm{CD}}^{N}(\bm{\theta^*}, \bm{\theta^*}; \bm{\phi}) = 0$, and considering that the ODE solver $\Phi(\cdots; \bm{\bm{\phi}})$ has the local discretisation error, Theorem 1 in~\cite{song2023consistency} shows that the consistency distillation error is bounded:

\begin{equation} \label{eq:dis_error}
    \mathop{\textnormal{sup}}\limits_{n, \vx}\Vert \vf_{\bm{\theta^*}}(\vx, t_{n})- \vf(\vx, t_n; \bm{\phi})  \Vert_2 = \mathcal{O} \left((\Delta t)^{p}\right),
\end{equation}
with $\Delta t$ and $p$ as defined in~\cite{song2023consistency}.

Besides the distillation strategy that needs an existing score model, \citet{song2023consistency} also introduced a way to train without any pre-trained models called Consistency Training (CT). 
In this paper, our primary focus is on the CD objective.

\subsection{Latent Consistency Model}

The Latent Consistency Model successfully extends consistency models~\cite{song2023consistency} to the image latent space corresponding with an auto-encoder~\cite{rombach2022high}. To leverage the advantages of CMs, the authors proposed Latent Consistency Distillation, where the consistency function with condition information $\vc$ can be defined as:
\begin{equation}
        \bm{f}(\bm{z}_t, \vc, t) \mapsto \bm{z}_0,
\end{equation}
which can directly predict \textit{the solution of PF-ODE} for $t=0$. To leverage the knowledge of pretrained text-to-image models effectively, LCMs introduce the \textit{parameterisation} method when parameterizing the consistency model with $\epsilon$-prediction teacher model:
\begin{equation} \label{eq:cf_lcm}
    \tilde{\vf}_{\bm{\theta}} (\vz_t, \vc,  t) = c_{\text{skip}}(t) \vz_t + c_{\text{out}}(t)\left( \frac{\vz_t - \sigma_t \hat\eps_{\bm{\theta}}(\vz_t,  \vc, t)}{\alpha_t} \right),
\end{equation}
where the boundary condition satisfies $c_{\text{skip}}(0)=1$ and $c_{\text{out}}(0)=0$, $\alpha_t$ and $\sigma_t$ specify the noise schedule in the perturbation kernels. It's worth noting that the above function, with appropriate parameterisation, is compatible with different prediction methods (\textit{e.g.}, $x$-prediction, $\epsilon$-prediction in~\cite{ho2020denoising}, $v$-prediction in~\cite{salimans2022progressive}).

To accelerate distillation training, \citet{luo2023latent} proposed the \textit{skipping-step} method to ensure self-consistency between timesteps $k$-steps apart, $t_{n+k} \rightarrow t_n$, instead of the original adjacent timesteps, $t_{n+1} \rightarrow t_n$. Additionally, \citet{luo2023latent} further implemented one-stage guided distillation~\cite{meng2023distillation} by solving the augmented PF-ODE. Then, the solution of $t_n$ from $t_{n+k}$ can be approximated with DDIM~\cite{song2020denoising} or DPM-Solver~\cite{lu2022dpm,lu2022dpm++}:
\begin{equation} \label{eq:lcm_solver}
\begin{split}
    \hat{\vz}_{t_n}^{\phi, \omega} = (1+\omega) &\Phi(\vz_{t_{n+k}}, t_{n+k}, t_n, \vc; \bm{\phi}) \\
    &- \omega \Phi(\vz_{t_{n+k}}, t_{n+k}, t_n, \varnothing ; \bm{\phi}),
\end{split}
\end{equation}
where $\Phi$ represents the update function of the PF-ODE solver with the pre-trained parameter $\phi$, and $\omega$ is the classifier-free guidance weight sampled from $\omega \in [\omega_{\text{min}}, \omega_{\text{max}}]$.

The objective function of LCM is derived from the self-consistency property, similar to the consistency distillation loss in~\cite{song2023consistency}:
\begin{equation} \label{eq:lcm_loss1}
\begin{aligned} 
 \mathcal{L}_{\rm{LCM}} (\bm{\theta},&\bm{\theta^-}; \bm{\Phi}) := 
    \mathbb{E}_{\vz, \vc, \omega, n} \bigg[ \\
    & d \left( \tilde\vf_{\bm{\theta}} (\vz_{t_{n+k}}, \vc,  t_{n+k}) ; \tilde\vf_{\bm{\theta}^-} (\hat{\vz}_{t_n}^{\Phi, \omega}, \vc,  t_n) \right) \bigg],
\end{aligned}
\end{equation}
where $d$ represents the distance between samples, with a common choice being the $\ell_2$ distance.

\section{Trajectory Consistency Distillation} \label{sec:methods}

\subsection{Background}
Given the $\epsilon$-prediction model $\epsilon_{\bm{\theta}}(\cdot)$, \citet{luo2023latent} used DDIM~\cite{song2020denoising} to re-parameterize the consistency function in~\cite{song2023consistency} but still followed the skip connection form as in~\cref{eq:cf_lcm} with boundary conditions $c_{\text{skip}}(0)=1$ and $c_{\text{out}}(0)=0$.
However, such a parameterization process includes a discretization error caused by the finite difference approximation of an explicit numerical method~\cite{song2020denoising} with the spacing $t\rightarrow 0$. Besides, in practical experiments with LCM released training codes~\footnote{\href{https://github.com/luosiallen/latent-consistency-model/blob/a9ad79587cc8bd1e404ccd1a3056a3da969b2f62/LCM\_Training\_Script/consistency\_distillation/train\_lcm\_distill\_sdxl\_wds.py\#L315}{https://github.com/luosiallen/latent-consistency-model}}, we noticed the implementation of
preconditions $c_{\text{skip}}$ and $c_{\text{out}}$ only matches the \textit{boundary condition} but stray from the initial definition in~\cite{karras2022elucidating}.

In order to better fit the formulation in LDM~\cite{rombach2022high} and SDXL~\cite{podell2023sdxl}, we first dispose of the skip-connection form with preconditions $c_{\text{skip}}$ \& $c_{\text{out}}$ and rewrite~\cref{eq:cf_lcm} as:
\begin{equation} \label{eq:cf_lcm_ddim}
\begin{split}
\tilde{\vf}_{\bm{\theta}} (\vx_t, t) &= \frac{\vx_t-\sigma_t\hat\eps_{\bm{\theta}}(\vx_t,t)}{\alpha_t} \\
&= \frac{1}{\alpha_t} \vx_t - \frac{\sigma_t}{\alpha_t} \hat\eps_{\bm{\theta}} (\hat{\vx}_{t}, t) \\
&=\frac{\alpha_0}{\alpha_t}\vx_t - \alpha_0 \left(\frac{\sigma_t}{\alpha_t} - \frac{\sigma_0}{\alpha_0}\right) \hat\eps_{\bm{\theta}} (\hat{\vx}_{t}, t),
\end{split}
\end{equation}
which is in the form of DDIM~\cite{song2020denoising}. As we mentioned before, this function also includes an approximation error with huge spacing.

~\citet{lu2022dpm} pointed out~\cref{eq:cf_lcm_ddim} is actually an approximated case for the exact solution of PF-ODEs with \emph{semi-linear} structure with an exponential integrator in terms of $\bm{\epsilon}_{\bm{\theta}}$:
\begin{equation}
\label{eq:exponential_integrators_raw}
    \hat\vx_s = \frac{\alpha_s}{\alpha_t} \vx_t + \alpha_s \int_{\lambda_t}^{\lambda_s} e^{-\lambda} \hat{\bm{\epsilon}}_{\bm{\theta}} (\hat{\vx}_{\lambda}, \lambda) \text{d} \lambda,
\end{equation}
and analyzed the corresponding error associated with $\lambda_t$ and $\lambda_s$, characterized as $\mathcal{O}((\lambda_s - \lambda_t)^k)$ with $k$-th Taylor expansion. In this context, $\bm{\epsilon}_{\bm{\theta}}$ is a trainable network with parameter $\bm{\theta}$, and $\lambda_t:=\log(\alpha_t / \sigma_t)$ is the log-SNR. We note that when using the same order $k$ of Taylor expansion, the error decreases as the interval between $s$ and $t$ reduces. Another point worth noting is that shortening the time intervals makes the boundedness assumption in~\cite{lu2022dpm} easier to meet, which ensures the error correspondingly decreases as the order $k$ increases.
We would like to highlight that such an mentioned error is the discretization error arising during the \textit{parameterisation process} in ~\cref{eq:cf_lcm}.
To further improve the guided sampling quality, another exact solution of PF-ODEs with $\vx_{\bm{\theta}}$ is proposed in~\cite{lu2022dpm++}:
\begin{equation}
\label{eq:exponential_integrators_xraw}
    \hat{\vx}_s = \frac{\sigma_s}{\sigma_t} \vx_t + \sigma_s \int_{\lambda_t}^{\lambda_s} e^{\lambda} \hat{\vx}_{\bm{\theta}}(\hat{\vx}_{\lambda}, \lambda)   \text{d} \lambda,
\end{equation}
which leads to better guided sampling performance and shares analogous properties with~\cref{eq:exponential_integrators_raw}. However, given that the consistency function is devoid of any guidance sampling, these two equations are equally valid views and can be used interchangeably~\cite{kingma2021variational}. In this work, we opt for the $\vx_{\bm{\theta}}$ representation due to its greater intuitiveness,
despite the fact that $\bm{\epsilon}_{\bm{\theta}}$ is mostly used for parameterization in practice.

Furthermore,~\cref{eq:dis_error} also reveals that in the distillation process, the upper bound of the CD loss is not only related to the number of steps $\Delta t$ derived by solvers, but is also associated with the time interval $t \rightarrow 0$ spanned in the consistency function.

\begin{figure}[t!]
    \centering
    \includegraphics[width=\linewidth]{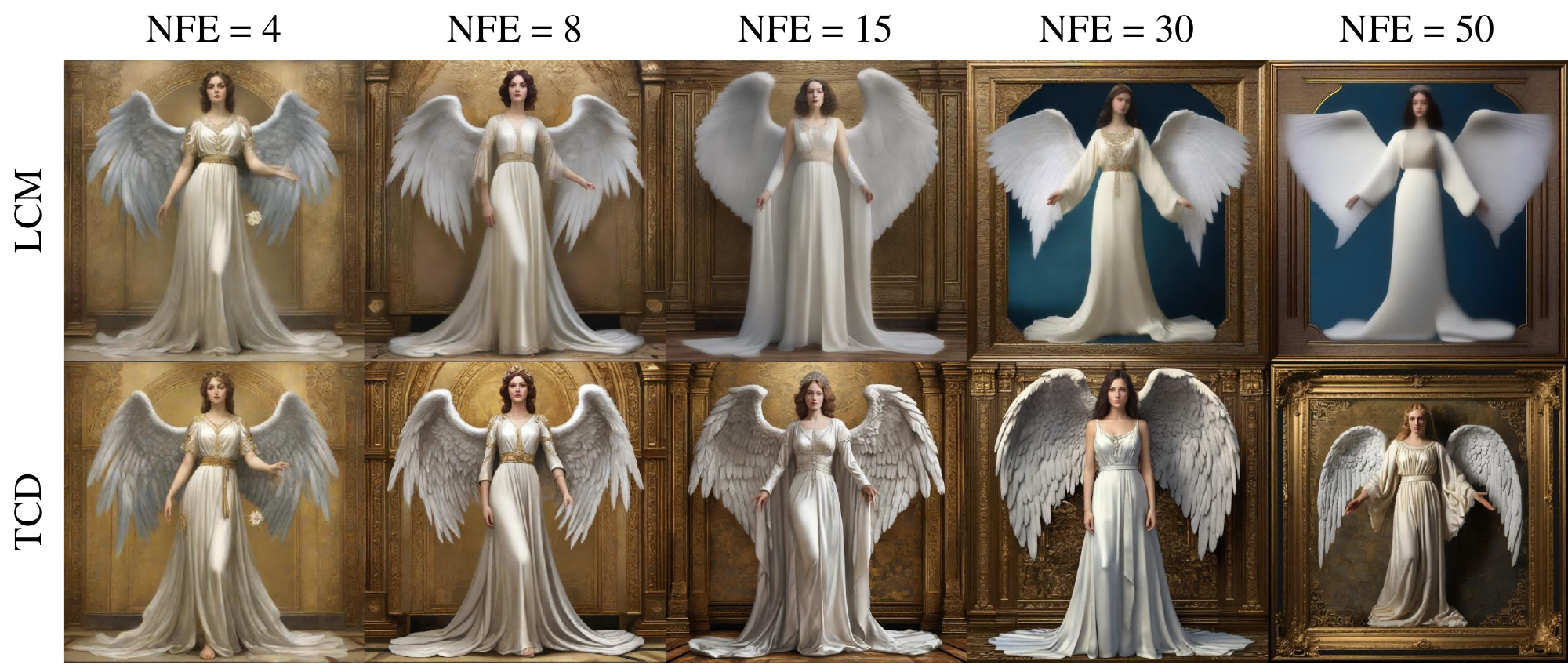}
    \caption{Synthesis results across various NFEs. Due to accumulated errors in multistep sampling, LCM experiences a loss of image detail, leading to a degradation in performance, whereas TCD addresses this issue. Additional samples are available in~\cref{app:diff_nfes}.}
    \label{fig:diff_nfes}
\end{figure}

\subsection{Trajectory Consistency Function}

Such errors as mentioned lead us to seek a better form for the consistency function by broadening the endpoint from the origin to a midpoint $s$ between $0$ and $t$ within the semi-linear structure.

Recall~\cref{eq:generalform} and the definition of the consistency function in~\cite{song2023consistency}: $\vf (\vx_{t}, t) \mapsto \vx $, and the definition of the \emph{self-consistency} property of the consistency function ~\cref{eq:consistency_prop}: \emph{The outputs of the consistency function are consistent for arbitrary pairs that belong to the same PF-ODE trajectory}. The goal of consistency distillation is to estimate this consistency function and ensure the self-consistency property.

\paragraph{Definition.}

We introduce the Trajectory Consistency Function (TCF) to re-parameter the output of $\vx_{\bm{\theta}}(\cdot)$ to $\vx_s$:
\begin{equation} \label{eq:tcf}
    \vf^{\rightarrow s} (\vx_{t}, t)  \mapsto \vx_s.
\end{equation}

Regarding the selection of $s$, our objective is to ensure that $s$ conforms to the minimum error across various discrete situations; thus, $s$ is designated as an arbitrary integer within the range $[0, t]$.

And TCF also supports \emph{self-consistency}. TCF remains consistent for arbitrary sets $(\vx_{t}, t)$ with the given $s$ that belong to the same PF ODE trajectory:
\begin{equation}
    \vf ^{\rightarrow s} (\vx_{t}, t) = \vf ^{\rightarrow s} (\vx_{t'}, t') \quad \forall t,t',s \in [0, T].
\end{equation}

\paragraph{Parameterisation.}
The \emph{semi-linear} structure with the exponential integrators form~\cite{lu2022dpm, lu2022dpm++, zhang2022fast} can be easily used to parameterise the trajectory consistency function,
\begin{equation}
\label{eq:exponential_integrators}
    \text{TCF:} \quad  \vf^{\rightarrow s}_{\bm{\theta}} (\vx_{t}, t)  = \frac{\sigma_s}{\sigma_t} \vx_t + \sigma_s \int_{\lambda_t}^{\lambda_s} e^{\lambda} \hat{\vx}_{\bm{\theta}} (\hat{\vx}_{\lambda}, \lambda) \text{d} \lambda.
\end{equation}

For $k \geqslant 1$, one can take the ($k$-1)-th Taylor expansion at $\lambda_t$ for $\vx_{\bm{\theta}}$ w.r.t $\lambda \in [\lambda_{s}, \lambda_{t}]$, we have:
\begin{equation}
\label{eq:taylor_f}
\begin{split}
    \vf^{\rightarrow s}_{\bm{\theta}} & (\vx_{t}, t) = \frac{\sigma_s}{\sigma_t}  \vx_t  + \\
    & \sigma_s \sum_{n=0}^{k-1} \hat\vx_{\bm{\theta}}^{(n)}(\hat{\vx}_{\lambda_t}, \lambda_t) \int_{\lambda_t}^{\lambda_s} e^{\lambda} \frac{(\lambda-\lambda_t)^n}{n!} \text{d} \lambda + \mathcal{O}(h^{k+1}),
\end{split}
\end{equation}
where $h=\lambda_s - \lambda_t$ and $\vx_{\bm{\theta}}^{(n)}(\cdot, \cdot)$ is the $n$-th order total derivatives of $\vx_{\bm{\theta}}$ w.r.t $\lambda$.

Here, we consider 1st-order and 2nd-order estimations, omitting high-order terms $\mathcal{O}(h^{k+1})$ in equation expression. 
\begin{equation}
\label{eq:tcd-1}
    \text{TCF(1):} \quad \tilde{\vf}^{\rightarrow s}_{\bm{\theta}} (\vx_{t}, t) = \frac{\sigma_s}{\sigma_t} \vx_t - \alpha_s (e^{-h} - 1) \hat{\vx}_{\bm{\theta}} (\vx_t, t).
\end{equation}

For the $2$nd-order expansion, we can write the trajectory consistency function as:
\begin{equation}
\label{eq:tcd-2}
\begin{split}
\text{TCF(2):} \quad \tilde{\vf}^{\rightarrow s}_{\bm{\theta}} (\vx_{t}, &t) = \frac{\sigma_s}{\sigma_t} \vx_t - \alpha_s (e^{-h} -1 ) \\
& \left( (1-\frac{1}{2r}) \hat\vx_{\theta} (\vx_{t}, t) + \frac{1}{2r}\hat\vx_\theta (\hat\vx_u, u) \right),
\end{split}
\end{equation}
where $u$ is the mid-timestep w.r.t. $t>u>s$ and $r := (\lambda_u - \lambda_t) / h$.

According to Theorem 3.2 in~\cite{lu2022dpm}, the upper error bounds for TCF parameterisation is the same as DPM-Solver-$k$. 
Let $\vf_{\bm{\theta}} ^{\rightarrow s} (\vx_{t}, t)$ be the ideal trajectory consistency function with exponential integrators form of the empirical PF-ODE, where $t$ and $s$ are the start timestep and end timestep along PF-ODE, we have the discretisation error regarding to the finite $k$-th order of TCF:
\begin{equation}
\label{eq:tcfde}
    \Vert \tilde{\vf}_{\bm{\theta}} ^{\rightarrow s} (\vx_{t}, t)- \vf_{\bm{\theta}}^{\rightarrow s} (\vx_{t}, t) \Vert_2 =  \mathcal{O}(h_{\Delta t}^{k+1}),
\end{equation}
where $h:= \lambda_s - \lambda_t$ and $k$ is the TCF order.
As identified in ~\cite{zhang2022fast, lu2022dpm}, employing semi-linearity in diffusion ODEs results in minimal error.

We also propose a modified network $F_{\bm{\theta}}$ with additional parameters for conditioning on $s$ to directly estimate \textit{the exponentially weighted integral} of $\vx_{\bm{\theta}}$ without omitting residual term when $k=1$ in~\cref{eq:taylor_f}:
\begin{equation}
\label{eq:tcd-s+}
     \text{TCF(S+):} \quad \tilde{\vf}_{\bm{\theta}} (\vx_{t}, t, s) =  \frac{\sigma_s}{\sigma_t}  \vx_t  - \alpha_s (e^{-h} - 1) \hat{F}_{\bm{\theta}} (\vx_t, t, s).
\end{equation}

In order to unify the representation of TCF, we abuse
\begin{equation}
    \vf(\vx_{t}, t, s) := \vf ^{\rightarrow s} (\vx_{t}, t),
\end{equation}
where $s$ can be a parameter either in the reparameterised process or the network.

We observe that TCF(S+) integrates the parameter $s$ into the network, yielding a formulation akin to the $G$ function in ~\cite{kim2023consistency}, albeit with a different coefficient. The underlying reason is that both instances employ a neural network to directly get the integral term from $t$ to $s$. The key distinction lies in the dependency of CTM's $G_{\theta}$ on the Euler Solver and the requirement of a $\vg_\theta$ for estimating $\mathbb{E}[\vx| \vx_t]$. Conversely, TCFs adopt a semi-linear structure for a better consistency function and use the finite difference approximation to parameterize the integral term.

\paragraph{Broaden Boundary Conditions} The boundary condition in the CM is limited to intervals extending from any start point on the solution trajectory to the origin.
And TCF is required to adhere to classical boundary conditions~\cite{song2020score}
\begin{equation}
  \vf (\vx_{0}, 0, 0) = \vx_0,  
\end{equation}
but also needs to meet a broaden boundary conditions
\begin{equation}
    \vf (\vx_{s}, s, s) = \vx_{s},
\end{equation}
to encompass a more comprehensive range of trajectory intervals with arbitrary sets.

It is evident that our parametrisation satisfy both the classical boundary conditions and the broaden boundary condition effortlessly.

\paragraph{Training.}
Considering $0 = t_1 < t_2 \cdots < t_N = T$ and given the one-step update function of a trained PF ODE solution $\Phi(\cdots; \bm{\phi})$ parameterised by $\bm{\phi}$, we can obtain an accurate estimation $\vx_{t_n}$ from $\vx_{t_{n+k}}$ by executing $k$ discretisation steps with $\Phi^{(k)}(\cdots; \bm{\phi})$,
\begin{equation} \label{eq:k_solver}
    \hat{\vx}_{t_n}^{\phi, k} = \Phi^{(k)}(\vx_{t_{n+k}}, t_{n+k}, t_n; \bm{\phi}).
\end{equation}

Thus, we could express the object of trajectory distillation in alignment with reconstruction, similar to the consistency distillation loss in~\cite{song2023consistency}:
\begin{equation} \label{eq:tcd_loss1}
\begin{aligned} 
 \mathcal{L}^{N}_{\rm{TCD}} (\bm{\theta},&\bm{\theta^-}; \bm{\phi}) := 
    \mathbb{E} [ \omega(t_n, t_m) \\
    &\Vert \vf_{\bm{\theta}} (\vx_{t_{n+k}}, t_{n+k}, t_m) - \vf_{\bm{\theta}^-} (\hat{\vx}_{t_n}^{\phi, k}, t_n, t_m) \Vert^2_2 ],
\end{aligned}
\end{equation}
where $n \sim \mathcal{U}[\![1, N-1]\!]$, $m \sim \mathcal{U}[\![1, n]\!]$, $\bm{\theta}^-$ can be either updated by EMA: $\bm{\theta}^- \leftarrow \texttt{sg}( \mu\bm{\theta}^-+(1-\mu)\bm{\theta})$ or stop the gradient without updating: $ \texttt{sg}(\bm{\theta})$, $\omega(\cdots)$ is a positive weighting function, and we find $\omega(t_n, t_m) \equiv 1$ performs well in our experiments. We also employ the skipping-step method proposed in~\cite{luo2023latent} to accelerate convergence. The detailed training process is outlined in~\cref{alg:tcd}. 

This objective function inherit the form in~\cite{luo2023latent, song2023consistency}, we expand it to a broaden condition according to the definition of TCF in \cref{eq:tcf} and $\vf_{\bm{\theta}}(\vx_t,t,s)$ still holds the form of $\vf^{\rightarrow s}_{\bm{\theta}} (\vx_{t}, t)$, which can be easily reparameterised and lead to a fast convergence. Such parameterisation of consistency function in semi-linear is pivotal for fast convergence and better results.

From Theorem 1 in~\cite{song2023consistency}, we could found trajectory consistency distillation with a shorter spacing can reduce the distillation error from $\mathcal{O} \left((\Delta t)^{p}\right) (t_n - t_0) $ to $\mathcal{O} \left((\Delta t)^{p}\right) (t_n - t_m)$.

It is observed that the TCF  not only leads the reduction in the parameterisation error bound as delineated in~\cref{eq:tcfde}, but also lowers the distillation error bound.

\subsection{Strategic Stochastic Sampling}

It is worth noting that once our TCF is trained, one can use common PF-ODE solvers like DDIM or SDE-DPM-Solver to sample high quality images. But it is not easy to control the stochastic strength with existing solvers; hence, we proposed Strategic Stochastic Sampling for better stochasticity control.

\paragraph{Background.}
Recalling the Multistep Consistency Sampling process in~\cref{alg:mcs}, each iteration of Multistep Consistency Sampling from $t$ to $s$ can be expressed as:
\begin{equation} \label{eq:mcs}
\begin{aligned}
    \text{Denoise:} \quad & \vx_{t \rightarrow 0} \leftarrow \vf_{\bm{\theta^*}}({\vx}_{t}, t), \\
    \text{Diffuse:} \quad & \hat{\vx}_{s} \leftarrow \alpha_{s} \vx_{t \rightarrow 0} + \sigma_{s} \vz,
\end{aligned}
\end{equation}
we found that the above process with parameterisation consistency function (\cref{eq:cf_lcm_ddim}) can be rewritten as a DDIM sampling equation ~\cite{song2020denoising} with $\eta = \sqrt{1 - \alpha_{s}^2}$ in~\cref{eq:generalddim}:

\begin{equation} \label{eq:mcs_ddim}
    \vx_{s} = \alpha_{s} \underbrace{\left(\frac{\vx_{t} - \sigma_{t} \epsilon_{\bm{\theta}}(\vx_{t})}{\alpha_{t}} \right)}_{\text{denoise}} + \underbrace{\sigma_{s} \vz}_{\text{diffuse}}.
\end{equation}

\cref{eq:mcs_ddim} leads us to review the general version of DDIM:
\begin{equation}
\label{eq:generalddim}
\begin{split}
    \vx_{s} &= \alpha_{s} {\left(\frac{\vx_t - \sigma_t \epsilon_{\bm{\theta}}(\vx_t)}{\alpha_t} \right)} \\
    &\quad \quad \quad \quad + \sqrt{1- \alpha_{s}^2 - \eta ^2} \epsilon_{\bm{\theta}}(\vx_t) + \eta \vz,
    \end{split}
\end{equation}
where occurs a parameter $\eta$ to control the stochasticity. And many related works~\cite{karras2022elucidating, song2020denoising, zhang2022fast} observed that few stochasticity can help reduce the sampling error, but high stochasticity will lead to poor quality. Such stochasticity during multistep sampling will obviously degrade the final output.

Inspired by the ideas of stochasticity control term from DDIM and SDE-DPM-Solvers~\cite{lu2022dpm++}, we propose Strategic Stochastic Sampling (SSS), which explicitly integrates the noise control term as an input parameter in the sampling process. Besides, our SSS could also be extended to different orders based on SDE-DPM-Solvers~\cite{lu2022dpm++}. Here we demonstrate the 1-st order SSS solver, which is based on~\cref{eq:generalddim}:

\begin{equation}\label{eq:controlled_para}
\begin{split}
    & \vx_{s} = \alpha_{s} \left(\frac{\vx_{t} - \sigma_{t} \epsilon_{\bm{\theta}}(\vx_{t})}{\alpha_{t}} \right) 
    + \alpha_{s} \frac{\sigma_{s^{'}}}{\alpha_{s^{'}}} \epsilon_{\bm{\theta}}(\vx_{t}) + \sqrt{1-\frac{\alpha_{s}^2}{\alpha_{s^{'}}^2}} \vz \\
    &= \frac{\alpha_{s}}{\alpha_{s^{'}}} \underbrace{\left(\alpha_{s^{'}} \frac{\vx_{t} - \sigma_{t} \epsilon_{\bm{\theta}}(\vx_{t})}{\alpha_{t}} + \sigma_{s^{'}} \epsilon_{\bm{\theta}}(\vx_{t}) \right)}_{\text{predicted} \; \vx_{s^{'}} = \vf(\vx_t, t, s^{'})}
    + \underbrace{\sqrt{1-\frac{\alpha_{s}^2}{\alpha_{s^{'}}^2}} \vz}_{\text{controllable noise}},
\end{split}
\end{equation}
instead of $\eta$ in~\cref{eq:generalddim}, we introduced an \emph{stochastic parameter} $\gamma$ that allows the control of the stochasticity in every step. 
As previously noted, the SSS framework is derived from the parameterisation of TCF and possesses the capability to explicitly manage stochastic intensities. Consequently, we define $\gamma$ as the parameterisation of target timestep $s$, with the linear range between full stochasticity and determinacy, thus $\gamma = 1 - s' / s $ and $\eta = \sqrt{1-{\alpha_{s}^2} / {\alpha_{s^{'}}^2}}$.

\paragraph{Sampling Process.}
Similar to Multistep consistency sampling, we can simply decompose~\cref{eq:controlled_para} into the format of multi-step sampling, which is demonstrated in~\cref{alg:sss}:
\begin{align}
 \text{Denoise:} \quad & \vx_{t \rightarrow s'} \leftarrow \vf_{\bm{\theta^*}}^{\rightarrow s'}(\vx_t, t) \label{eq:sss_denoise}  \\
 \text{Diffuse:} \quad & \hat{\vx}_{s} \leftarrow \frac{\alpha_{s}}{\alpha_{s'}} \vx_{t \rightarrow s'}  + \sqrt{1- \frac{\alpha_{s}^2}{\alpha_{s'}^2}} \vz
\end{align}

Specifically, every sampling step in SSS includes the \textit{denoise sub-step} according to the ODE solver and the \textit{diffuse sub-step} based on Langevin SDE. In comparison with multistep consistency sampling, where the endpoint and noise level are fixed, SSS introduces the additional parameter $\gamma$ to control the destination point for the denoise step and allows for the adjustment of random noise level for the diffuse step, as detailed in~\cref{fig:sss} and~\cref{alg:sss}.

In \cite{kim2023consistency}, the authors proposed a unified sampling method and have delicately crafted a $\gamma$-sampler for its $G_{\theta}$ function to leverage its ability for “anytime to anytime” prediction. This sampler incorporates a $\gamma$ parameter to regulate the decoder $G_{\theta}$ target, yet further endows it with the properties of both EDM sampler and deterministic sampler within the Multistep Consistency Sampling protocol. Another key finding in~\cite{kim2023consistency} is the analysis of the error bound in $\gamma$-sampling $\mathcal{O}(\sum_{n=0}^{N-1}\sqrt{t_n - \sqrt{1-\gamma^2}t_{n+1}})$, which aligns with the previous practical observation~\cite{karras2022elucidating}: less stochastic will lead to a better sampling quality.

It is worth noting that SSS was specially designed to control the stochasticity in LCM sampling with the well-defined $\gamma$ drived from DDIM sampling. Additionally, its alternating sampling framework is grounded on multistep consistency sampling, similar to the $\gamma$ sampling in~\cite{kim2023consistency}.
From this perspective, we notice that SSS is also capable of changing the stochastic intensity during its sampling procedure and has the same error bound as stated in~\cite{kim2023consistency}. 
In another word, our method is also able to reduce the errors that arise from the $n$-th step to $\mathcal{O}(\sqrt{\tau_n - (1-\gamma) \tau_{(n+1)}})$ when $n \in [\![1, N-1]\!]$ and ultimately optimize the accumulated error. It is worth noting that in the process of sampling with TCF(1) and TCF(2), there is another discretization error $\mathcal{O}(h_{\Delta \tau}^{k+1})$ that cannot be eliminated, which is brought by the parameterisation of TCF as discussed in \cref{eq:tcfde}.

We would like to highlight that the parameterisation form of TCF makes it versatile for community samplers, and our SSS provides an additional method to explicitly control the stochastic level for LCM.
It is also worth noting that when $\gamma$ is low, the estimation error plays a more important role~\cite{lyu2023convergence}. Hence, the optimal value of $\gamma$ should be determined empirically, as we show in~\cref{sec:ablation}.

\begin{figure*}[ht!]
    \centering
    \includegraphics[width=\linewidth]{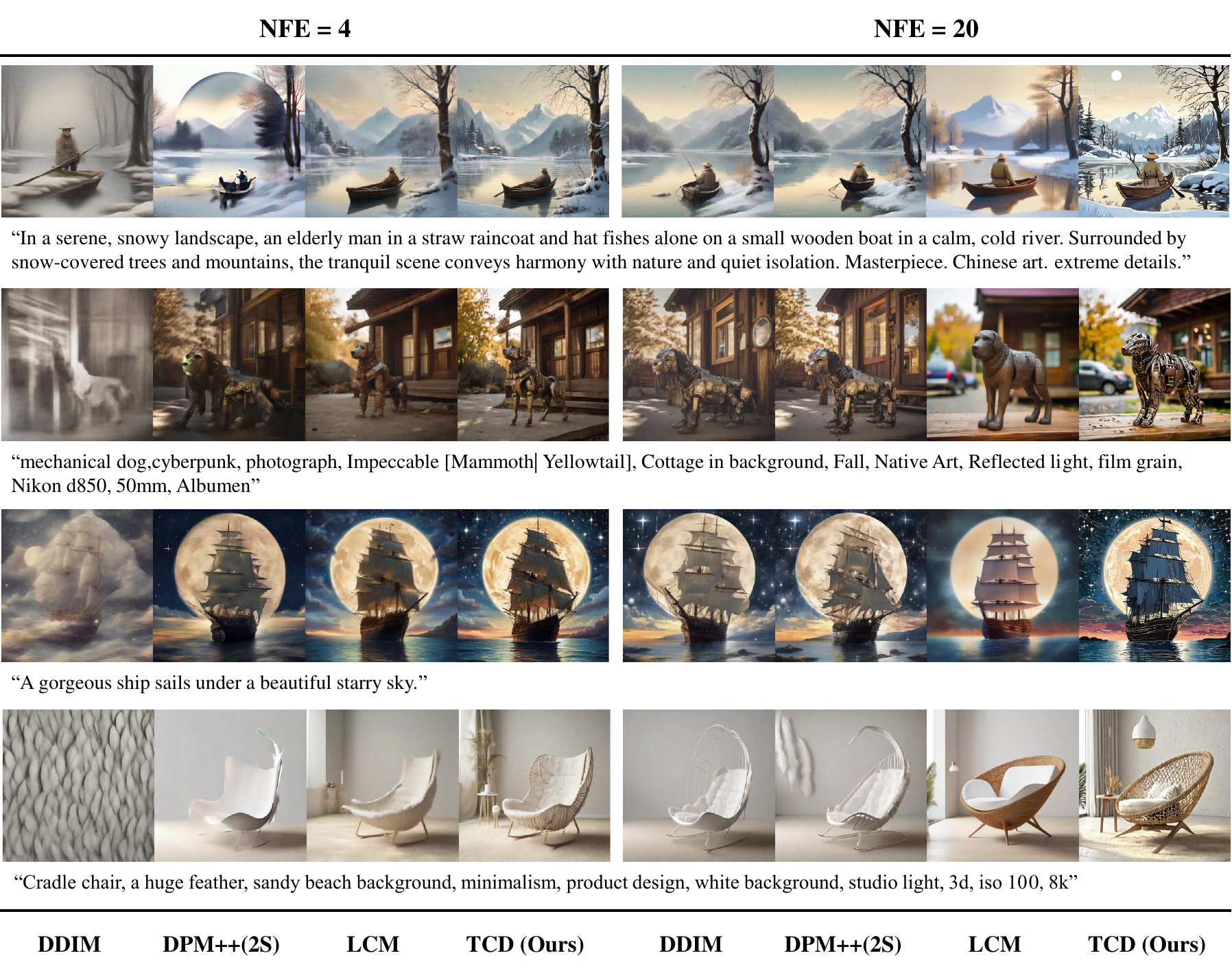}
    \caption{Qualitative comparison. For each prompt, images are generated using the same random seed for every model, without any cherry-picking. More results are provided in~\cref{app:more_compare}.}
    \label{fig:quality}
\end{figure*}
\begin{figure*}[ht!]
    \centering
    \includegraphics[width=\linewidth]{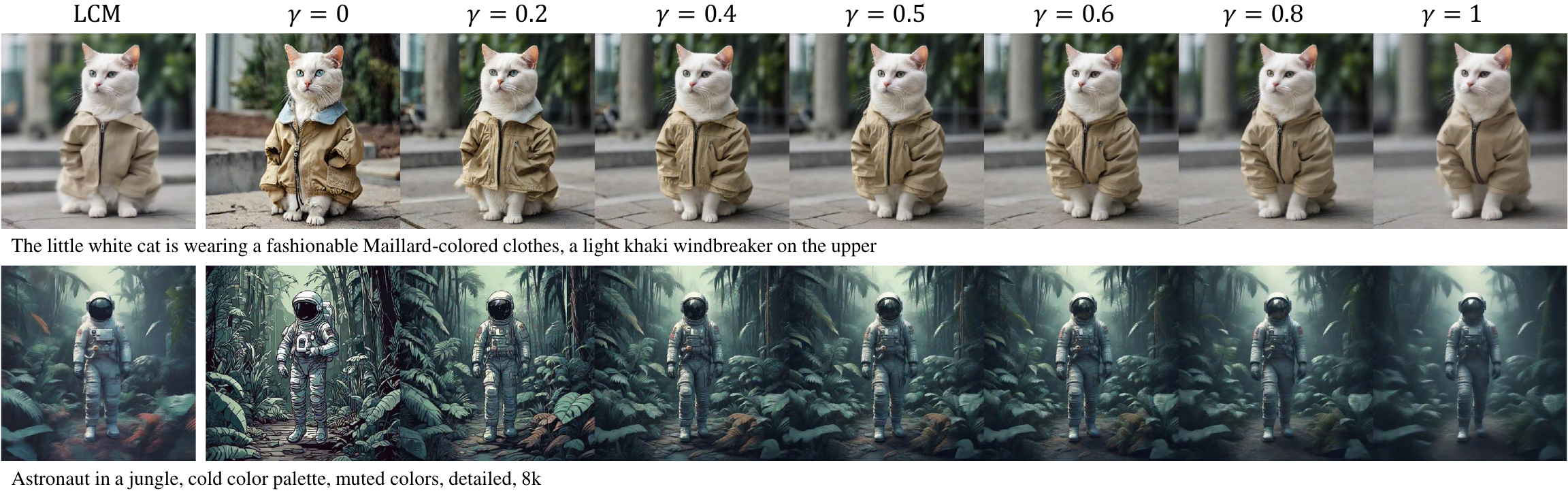}
    \caption{Qualitative effects of stochastic parameter $\gamma$ with same NFEs. Images under the same prompt but with different $\gamma$ applied during sampling. The leftmost image is sampled from LCM~\cite{luo2023lcm}. More samples can be found in~\cref{app:diff_gamma}.}
    \label{fig:diff_gamma_fig}
\end{figure*}
\begin{table*}[]
\caption{Quantitative comparison on the COCO validation set.}
\label{tab:compare}
\centering
\fontsize{5}{6.5}\selectfont
\resizebox{\linewidth}{!}{%
\begin{threeparttable}[b]
\begin{tabular}{l|cccc|cccc}
\hline
\multirow{2}{*}{METHOD} & \multicolumn{4}{c|}{FID $\downarrow$}                             & \multicolumn{4}{c}{Image Complexity Score $\uparrow$}                 \\ \cline{2-9} 
                        & 2 STEPS        & 4 STEPS        & 8 STEPS        & 20 STEPS       & 2 STEPS         & 4 STEPS         & 8 STEPS         & 20 STEPS        \\ \hline
Euler~\cite{karras2022elucidating}                   & 104.73         & 44.31          & 18.20          & 14.72          & 0.4251          & 0.3639          & 0.4151          & 0.4489          \\
DDIM~\cite{song2020denoising}                    & 105.98         & 44.86          & 17.62          & 13.60          & {\ul 0.4456}    & 0.3633          & 0.4148          & 0.4481          \\
DPM++(2S)~\cite{lu2022dpm++}               & 46.08          & 18.50          & \textbf{12.49} & \textbf{12.15} & 0.2876          & {\ul 0.4496}    & {\ul 0.4788}    & {\ul 0.4679}    \\
LCM~\cite{luo2023latent}                     & {\ul 16.15}    & {\ul 15.03}    & 16.93          & 18.13          & 0.4300          & 0.4364          & 0.4260          & 0.4057          \\
TCD (Ours)              & \textbf{14.66} & \textbf{12.68} & {\ul 13.64}    & {\ul 13.56}    & \textbf{0.4701} & \textbf{0.5095} & \textbf{0.5336} & \textbf{0.5563} \\ \hline \hline
\multirow{2}{*}{METHOD}               & \multicolumn{4}{c|}{ImageReward $\uparrow$}                       & \multicolumn{4}{c}{PickScore $\uparrow$}                              \\ \cline{2-9} 
                        & 2 STEPS        & 4 STEPS        & 8 STEPS        & 20 STEPS       & 2 STEPS         & 4 STEPS         & 8 STEPS         & 20 STEPS        \\ \hline
Euler~\cite{karras2022elucidating}                   & -227.77        & -189.41        & 12.59          & 65.05          & 16.75           & 18.71           & 21.32           & 22.21           \\
DDIM~\cite{song2020denoising}                    & -227.75        & -189.96        & 13.45          & 66.14          & 16.74           & 18.68           & 21.31           & 22.16           \\
DPM++(2S)~\cite{lu2022dpm++}               & -169.21        & -1.27          & {\ul 67.58}    & \textbf{75.8}  & 19.05           & 20.68           & 21.9            & {\ul 22.33}     \\
LCM~\cite{luo2023latent}                     & {\ul 18.78}    & {\ul 52.72}    & 55.16          & 49.32          & {\ul 21.49}     & {\ul 22.2}      & {\ul 22.32}     & 22.25           \\
TCD (Ours)              & \textbf{34.58} & \textbf{68.49} & \textbf{73.09} & {\ul 74.96}    & \textbf{21.51}  & \textbf{22.31}  & \textbf{22.5}   & \textbf{22.36}  \\ \hline
\end{tabular}
\begin{tablenotes}
    \item[ ]  The best scores are highlighted in \textbf{bold}, and the runner-ups are \underline{underlined}.
\end{tablenotes}
\end{threeparttable}
}
\end{table*}
\begin{figure}[t!]
    \centering
    \includegraphics[width=\linewidth]{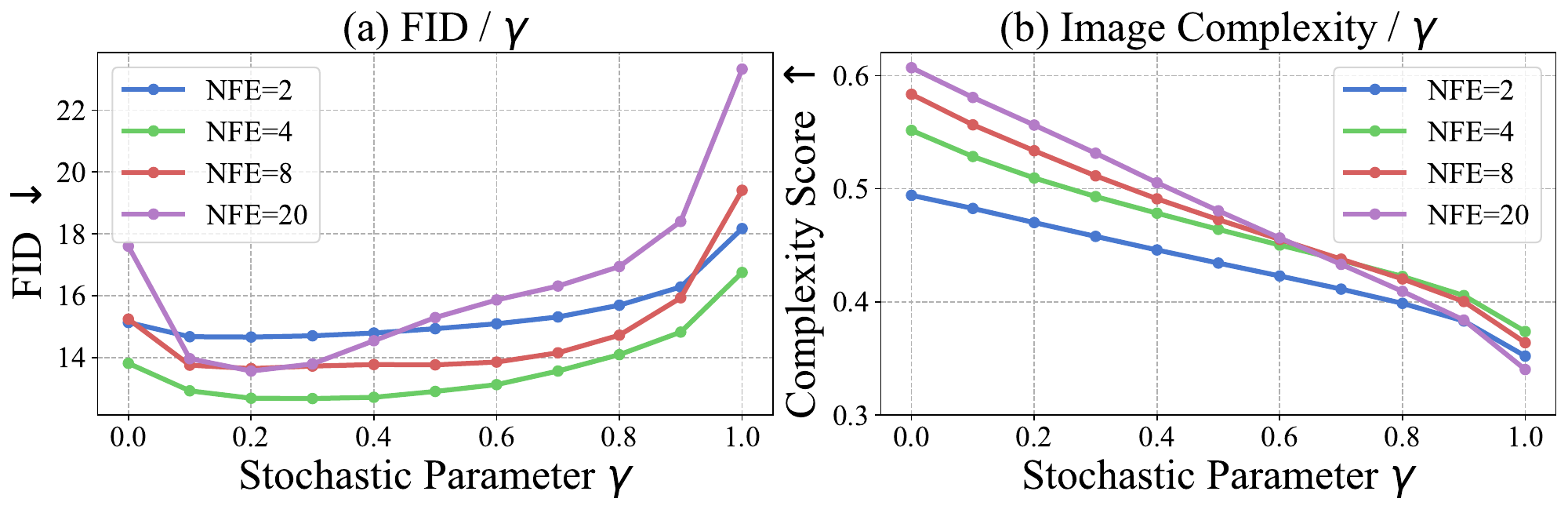}
    \caption{Quantitative ablation on different stochastic parameter $\gamma$.}
    \label{fig:diff_gamma_curve}
\end{figure}

\begin{table}[]
\caption{Quantitative ablation on the TCF parameterisation type.}
\label{tab:para_type}
\centering
\resizebox{\linewidth}{!}{%
\begin{tabular}{lcccc}
\toprule 
Para Type    & FID $\downarrow$ & IC Score $\uparrow$ & ImageReward $\uparrow$ & PickScore $\uparrow$ \\ \midrule
TCF(1)                        & 12.68            & 0.5095              & 68.49                  & 22.31                \\
TCF(2)                        & 13.35            & 0.5037              & 58.13                  & 22.07                \\
TCF(S+)                       & 13.03            & 0.4176              & 57.96                  & 22.01                \\ \bottomrule
\end{tabular}
}
\end{table}
\begin{figure*}[t!]
    \centering
    \includegraphics[width=\linewidth]{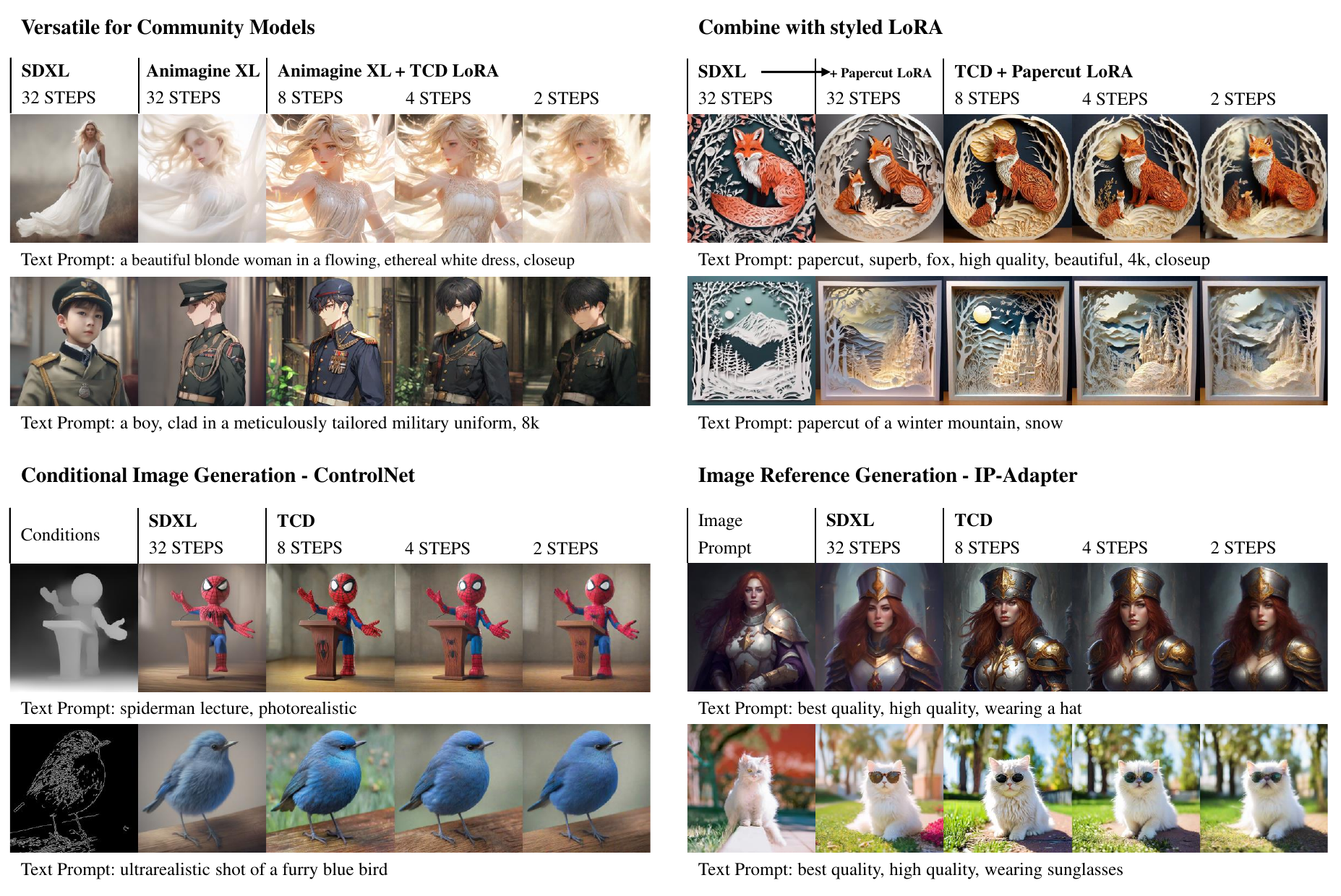}
    \caption{Versatility of TCD. Our TCD LoRA can be directly applied to a wide range of models, including custom community models, styled LoRA, ControlNet, and IP-Adapter, accelerating their generation in just a few steps with high quality.}
    \label{fig:versatility}
\end{figure*}

\subsection{Extension to Large Text Conditional Models}
Conditional models frequently outperform their unconditional counterparts and exhibit a wider range of use cases~\cite{bao2022conditional, dhariwal2021diffusion, ho2022classifier}. Particularly, text conditional models have recently garnered substantial attention, showcasing remarkable results~\cite{nichol2022glide, ramesh2022hierarchical, saharia2022photorealistic, podell2023sdxl}. The trajectory consistency function can be seamlessly integrated into conditional models by introducing an additional input, $\vc$, to accommodate conditioning information, such as text. This results in the transformation of the trajectory function to $\vf_{\rm{\theta}}(\vx_t, \vc, t, s)$, and the guided distillation method proposed by~\cite{meng2023distillation, luo2023latent} can be directly integrated into it, as detailed in~\cref{alg:guided_tcd}.

Trajectory consistency distillation can be considered as fine-tuning process, which directly occur on top of the pre-trained model. For scaling TCD to larger models (e.g., SDXL) with significantly reduced memory consumption, we incorporate Low-Rank Adaptation (LoRA)~\cite{hu2021lora}, a parameter-efficient fine-tuning method, into the distillation process. Additionally, the parameters of LoRA can be identified as a versatile acceleration module applicable to different fine-tuned models or LoRAs sharing the same base model without the need for additional training, aligning with the observations in~\cite{luo2023lcm}.

\section{Experiments}
\subsection{Experimental Setup}
\label{sec:exp_setup}
We selected SDXL~\cite{podell2023sdxl}, a widely recognized diffusion model, as our backbone.
By default, we employ TCF(1) as the parameterisation and set the stochastic parameter $\gamma$ as 0.2.
The influence of $\gamma$ and parameterisation type is left to be explored in the ablation studies (\cref{sec:ablation}).
For detailed implementation information, please refer to~\cref{app:imp_details}.

\subsection{Main Results}
\label{sec:main_exp}
To illustrate the effectiveness and superiority of our methods, we perform qualitative and quantitative comparisons with prior works, including Euler~\cite{karras2022elucidating}, efficient numerical ODE solvers like DDIM~\cite{song2020denoising} and DPM-Solver++(2S)~\cite{lu2022dpm++}, and a relevant work LCM~\cite{luo2023latent} as baseline.

\paragraph{Qualitative Results.}
As illustrated in ~\cref{fig:quality}, prior efficient numerical methods produce suboptimal images with 4 steps, while LCM can generate relatively better images. Our TCD further enhances visual quality. 
With an increased number of function evaluations (20 steps), the quality of samples generated by DDIM or DPM-Solver++(2S) improves rapidly. However, the improvement in LCM is not as noticeable, resulting in smoother and less detailed images due to the accumulated error in multistep sampling. 
In contrast, TCD addresses this flaw, producing more detailed images that surpass even the teacher model, \textit{e.g.,} SDXL with DDIM. 

\vspace{-8pt}
\paragraph{Quantitative Comparison.}
For the quantitative evaluation, we conducted zero-shot image generation using 5K captions from COCO2017 validation set. We employed the Frechet Inception Distance (FID) and the Image Complexity Score~\cite{feng2022ic9600} to assess the sample quality and visual complexity of generated images. Additionally, we used Image Reward~\cite{xu2023imagereward} and PickScore~\cite{kirstain2023pick} to quantify text-image alignment and human preference.
As depicted in~\cref{tab:compare}, TCD shows an improvement in the performance of LCM across various timesteps and metrics. In comparison to numerical ODE-Solver, TCD exhibits the ability to generate high-quality images in only 2 or 4 steps, surpassing them significantly. When increasing steps to 8 or 20, LCM's performance tends to decline, while TCD even outperforms its teacher in almost all metrics.

\subsection{Ablation Studies}
\label{sec:ablation}
\paragraph{Effects of Stochastic Parameter.} 
The effect of the stochastic parameter $\gamma$ is shown in~\cref{fig:diff_gamma_fig} and~\cref{fig:diff_gamma_curve}. As illustrated in~\cref{fig:diff_gamma_fig} and ~\cref{fig:diff_gamma_curve} (b), we find that with an increase in $\gamma$, the visual complexity and fineness of the image gradually improve.
Furthermore, our observation indicates that implementing a stochastic sampling process can reduce accumultaed errors. Notably, when the gamma value is set to 0, the inherent estimation error of the model becomes more pronounced, leading to a decrease in quality evaluated with FID, as shown in ~\cref{fig:diff_gamma_curve} (a).

\vspace{-10pt}
\paragraph{Parameterisation Type.}
We illustrate the impact of our proposed different parameterisation types in~\cref{tab:para_type} with 4 sampling steps. We observed that the instability of the high-order solver discussed in~\cite{lu2022dpm++} also exists in TCF(2), resulting in slightly inferior performance compared to TCF(1). Due to the introduction of additional parameters, TCF(S+) faces challenges in effectively distilling the teacher's information, leading to poorer performance. Note that while TCF(1) only requires 3000 training iterations to achieve satisfactory results in~\cref{tab:para_type}, TCF(S+) takes 43,000 iterations but still lags behind in performance.

\paragraph{Versatility of TCD.}
To assess the versatility of TCD, we extensively tested it on various models, including the popular community model Animagine XL V3~\footnote{\textit{Animagine}: https://civitai.com/models/260267/animagine-xl-v3}, styled LoRA Papercut XL~\footnote{\textit{Papercut}: https://civitai.com/models/122567/papercut-sdxl}, Depth ControlNet~\footnote{\textit{Depth ControlNet}: https://huggingface.co/diffusers/controlnet-depth-sdxl-1.0}, Canny ControlNet~\footnote{\textit{Canny ControlNet}: https://huggingface.co/diffusers/controlnet-canny-sdxl-1.0}, and IP-Adapter~\footnote{\textit{IP-Adapter}: https://github.com/tencent-ailab/IP-Adapter}. The results shown in~\cref{fig:versatility} imply that TCD can be directly applied to various models to accelerate image generation with high quality in only 2-8 steps. Additional samples can be found in~\cref{app:versatility_samples}.

\vspace{-8pt}
\section{Conclusion}
In this work, we introduce TCD, a improved latent consistency distillation method that reduces inherent errors present in latent consistency models, including TCF for training and SSS for sampling.  The TCF is proposed to diminish parameterisation and distillation errors and enable the model to track the trajectory along the PF ODE with a semi-linear structure. Furthermore, SSS is proposed to explicitly control the stochastic intensity and reduce accumulated errors by the bijective traversal. Remarkably, TCD outperforms LCM across all sampling steps and exhibits superior performance compared to numerical methods of teacher model. We believe that TCD can provide novel perspectives for fast and high-quality image generation, while certain characters of TCD also contribute valuable insights to downstream applications, \textit{e.g.}, enhanced details for super-resolution and a better intermediate manifold for editing.

\vspace{-8pt}
\paragraph{Limitations.} In our experiments, we observed instability in high-order TCF and poor convergence in TCF(S+). Further analysis is necessary to ascertain the stability of the high-order function and TCF(S+). Additionally, it is worth investigating an improved design to achieve fewer steps generation, \textit{e.g.}, single step.


\bibliography{references}
\bibliographystyle{icml2024}

\newpage
\appendix
\onecolumn
\section{Discussion between CTM~\citep{kim2023consistency} and TCD}~\label{ctmd}

A recent work, CTM~\cite{kim2023consistency}, proposed a unified framework for CM and DM solvers, with the idea of “anytime to anytime” prediction, theoretically substantiated and facilitated through a $G_\theta$ function with $\gamma$-sampling for assessing both the integrand and the integral of PF-ODE.
Conversely, TCD focused on improving the performance of LCM by leveraging a semi-linear structure with a narrower time spacing in the consistency function. While these two studies share intersections in  their approaches, such as the use of varying time intervals from $t$ to $s$, they also exhibit distinct differences.
In this section, we delve into a discussion on CTM and TCD, exploring their various aspects.

To initiate our comparison, we first recall two definitions: explicit numerical method and implicit numerical method for ODEs~\cite{atkinson2009numerical,suli2010numerical}:

\begin{quote}
A general explicit method can be:
\begin{equation}
y_{n+1} = y_n + h f(y_n,t_n), \,\, n = 0,\dots,N-1,
\end{equation}
where $f(y_n,t_n)$ is a continuous function and $h$ is the interval between $t_n$ and $t_{n+1}$. In an explicit method, $y_n$ is already calculated, and $y_{n+1}$ can be given directly.

Similarly, the implicit method, which is also said to be~\textit{absolutely stable}, can be expressed in the form:
\begin{equation}
y_{n+1} = y_n + h f(y_{n+1},t_{n+1}), \,\, n = 0,\dots,N-1,
\end{equation}
which requires the solution of an implicit equation $f(y_{n+1},t_{n+1})$ in order to determine $y_{n+1}$.

\end{quote}
A representative example of an explicit method is the Euler Method, while the Backward Euler Method is an implicit method.
Informally, score-based sampling with time-discretized numerical integral solvers can be regarded as an explicit method, whereas the distillation model that directly estimates the integral of ODEs is an implicit method~\cite{kim2023consistency}.

In order to unify CMs and DMs by assessing both the infinitesimal jump and long jump, CTM~\citep{kim2023consistency} proposed a unified decoder $G$ with an ``anytime-to-anytime" idea, where $G$ is the solution of the PF-ODE with an integrator term. Furthermore, CTM used a trainable $g_\theta$ to parameterize the integrator term and made it access the infinitesimal jump inspired by the Euler Solver. In order to train such a $G_{\theta}$ function, instead of directly forcing $G_{\theta}(\vx_t,t,s) \approx \texttt{Solver}(\vx_t,t,s;\phi)$, CTM proposed a soft consistency matching $G_{\theta} (\vx_t,t,s) \approx G_{\text{sg}(\theta)}(\texttt{Solver}(\vx_t,t,u;\phi),u,s)$ that supports both $u=s$ and $u=t-\Delta t$. It can be observed that this loss is similar to the distillation loss for consistency model~\cite{song2020denoising}. Additionally, CTM has also introduced DSM loss and GAN loss to optimize the final performance of the model. It's evident that the adversarial loss plays an important role in reaching a better FID. To further leverage the implicit method with a well-trained $g_\theta$ function, CTM proposed $\gamma$-sampling, which integrates the properties of both the EDM sampler and the deterministic ODE sampler, and falls within the protocol of multistep consistency sampling, i.e., iterative denoising and diffusing. If we were to apply the informal analogy of explicit and implicit methods, CTM proposed a distillation-based \textit{implicit method} with $G_{\theta}$ to address the problems in time-discretized explicit solvers.

In contrast, TCD aims to tackle the challenge posed by the imprecise parameterisation of the consistency function~\cite{song2023consistency} in LCM~\cite{luo2023latent} and improve the generation quality on text-to-image task.
Our method strictly adheres to the definitions of the consistency model, enabling the direct adoption of characteristics and algorithms inherent to consistency models. We first observed that several errors in the distillation process are related to the time interval $t \rightarrow s$ of the consistency function. Thus, we leverage the semi-linear structure with exponential integrators of the PF-ODE for parameterization, which also supports a shorter interval (i.e., moderately moving the upper limit $s$ away from $0$). Such concept of transitioning from $t$ to $s$ overlaps with CTM's ``anytime to anytime" idea but ours is from different view. Moreover, this semi-linear structure supports $\vx_0$, $\vv$, and $\epsilon$ parameterisation, allowing for efficient distillation from large-scale diffusion models.
During training, we simply follow all the definitions and targets of Consistency Distillation and equip $\mathbb{E} [ \omega(t_n, t_m)
\Vert \vf_{\bm{\theta}} (\vx_{t_{n+k}}, t_{n+k}, t_m) - \vf_{\bm{\theta}^-} (\hat{\vx}_{t_n}^{\phi, k}, t_n, t_m) \Vert^2_2 ]$ as the loss while $\hat{\vx}_{t_n}^{\phi, k}$ is from a solver by 1 NFE with parameter $\phi$ and $k \geq 1$. As observed in~\cite{sauer2023adversarial}, GAN loss tends to result in poorer mode coverage, hence we only employ the distillation loss used in~\cite{song2023consistency, luo2023latent} to ensure the diversity of generated images.
The sampling process of TCD remains \textit{explicit} as the TCF can be easily parameterized by $\epsilon$, allowing it to directly utilize existing explicit samplers. This means that once trained, TCF functions as a plug-and-play model, eliminating the need for designing additional samplers. Recognizing Multistep Consistency Sampling with parameterisation
consistency function in LCM as a fixed stochasticity DDIM solver~(\cref{eq:mcs_ddim}), we introduced SSS to achieve a flexible stochastic intensity control. 
Since CTM~\cite{kim2023consistency} proposed a unified framework for CMs and DMs and our SSS is directly based on multistep sampling in CMs, it demonstrates theoretical similarity to the $\gamma$-sampling used in CTM~\cite{kim2023consistency}. And the theoretical insights provided by CTM~\cite{kim2023consistency} in sampling errors can also support our conjectures, e.g., injecting more noise will lead to degraded quality.

In summary, CTM has four key contributions: 1. A single model inspired by Euler Solver that attains all the advantages of score-based models and distillation models ; 2. $\gamma$-sampling for flexible sampling; 3. Soft matching for efficient and stabilized training; 4. The use of auxiliary losses for better student training. Meanwhile, TCD is an enhancement of LCM~\cite{luo2023latent}, adhering to the definition of consistency models with the following features: 1. Better parameterization of \textit{consistency functions} with a semi-linear structure~\cite{lu2022dpm, lu2022dpm++, zhang2022fast}; 2. Maintains the \textit{explicit} sampling method and is versatile for existing solvers; 3. Employs only consistency loss with \textit{skipping-steps} to achieve self-consistency and avoid mode collapse. Moreover, TCD aims to improve the quality of general generation tasks in fewer steps, instead of achieving SOTA for single-step generation for specific tasks.


\section{Algorithmic Details}

\subsection{Algorithm Details of Trajectory Consistency Distillation}
We provided a detailed algorithm for trajectory consistency distillation during training as~\cref{alg:tcd} and a guided version for conditional models as~\cref{alg:guided_tcd}.

\begin{minipage}[t]{0.5\textwidth}
\begin{algorithm}[H]
   \caption{Trajectory Consistency Distillation}
   \label{alg:tcd}
   \setstretch{1.009}
\begin{algorithmic}
   \STATE {\bfseries Input:} dataset $\mathcal{D}$, initial model parameter $\bm{\theta}$, learning rate $\eta$, the update function of ODE solver $\Phi(\cdots; \bm{\phi})$, EMA rate $\mu$, noise schedule $\alpha_t$, $\sigma_t$, skipping interval $k$ \\
   $\bm{\theta}^- \leftarrow \bm{\theta}$
   \REPEAT
   \STATE Sample $\vx \sim \mathcal{D}$
   \STATE Sample $n \sim \mathcal{U}[\![1, N-k]\!]$, $m \sim \mathcal{U}[\![1, n]\!]$
   \STATE Sample $\vx_{t_{n+k}} \sim \mathcal{N}(\alpha_{t_{n+k}} \vx, \sigma_{t_{n+k}}^2 \bm{I})$
   \STATE $\hat{\vx}_{t_n}^{\phi, k} \leftarrow \Phi^{(k)}(\vx_{t_{n+k}}, t_{n+k}, t_n; \bm{\phi})$ 
   \STATE
   \STATE $\mathcal{L} (\bm{\theta}, \bm{\theta^-}; \bm{\phi}) \leftarrow$ \\
   $\Vert \vf_{\bm{\theta}} (\vx_{t_{n+k}}, t_{n+k}, t_m)-\vf_{\bm{\theta}^-} (\hat{\vx}_{t_n}^{\phi, k}, t_n, t_m) \Vert^2_2$
   \STATE $\bm{\theta} \leftarrow \bm{\theta} - \eta \nabla_{\bm{\theta}} \mathcal{L} (\bm{\theta}, \bm{\theta^-}; \bm{\phi})$
   \STATE $\bm{\theta^-} \leftarrow \texttt{sg} (\mu\bm{\theta}^-+(1-\mu)\bm{\theta})$
   \UNTIL{convergence}
\end{algorithmic}
\end{algorithm}
\end{minipage}
\begin{minipage}[t]{0.5\textwidth}
\begin{algorithm}[H]
   \caption{Guided Trajectory Consistency Distillation}
   \label{alg:guided_tcd}
\begin{algorithmic}
   \STATE {\bfseries Input:} dataset $\mathcal{D}$, initial model parameter $\bm{\theta}$, learning rate $\eta$, the update function of ODE solver $\Phi(\cdots; \bm{\phi})$, EMA rate $\mu$, noise schedule $\alpha_t$, $\sigma_t$, skipping interval $k$ \\
   $\bm{\theta}^- \leftarrow \bm{\theta}$
   \REPEAT
   \STATE Sample $(\vx, \vc) \sim \mathcal{D}$, $\omega \sim \mathcal{U}[\omega_\text{min}, \omega_\text{max}]$
   \STATE Sample $n \sim \mathcal{U}[\![1, N-k]\!]$, $m \sim \mathcal{U}[\![1, n]\!]$
   \STATE Sample $\vx_{t_{n+k}} \sim \mathcal{N}(\alpha_{t_{n+k}} \vx, \sigma_{t_{n+k}}^2 \bm{I})$
   \STATE $\hat{\vx}_{t_n}^{\phi, k} \leftarrow (1+\omega)\Phi^{(k)}(\vx_{t_{n+k}}, t_{n+k}, t_n, \vc; \bm{\phi}) -$ \\
   $\omega \Phi^{(k)}(\vx_{t_{n+k}}, t_{n+k}, t_n, \varnothing; \bm{\phi})$
   \STATE $\mathcal{L} (\bm{\theta}, \bm{\theta^-}; \bm{\phi}) \leftarrow$ \\
   $\Vert \vf_{\bm{\theta}} (\vx_{t_{n+k}}, \vc, t_{n+k}, t_m)-\vf_{\bm{\theta}^-} (\hat{\vx}_{t_n}^{\phi, k}, \vc, t_n, t_m) \Vert^2_2$
   \STATE $\bm{\theta} \leftarrow \bm{\theta} - \eta \nabla_{\bm{\theta}} \mathcal{L} (\bm{\theta}, \bm{\theta^-}; \bm{\phi})$
   \STATE $\bm{\theta^-} \leftarrow \texttt{sg} (\mu\bm{\theta}^-+(1-\mu)\bm{\theta})$
   \UNTIL{convergence}
\end{algorithmic}
\end{algorithm}
\end{minipage}

\subsection{The Difference between Multistep Consistency Sampling and Strategic Stochastic Sampling}
For simplicity, we consider VP SDEs in this paper.
Thus, the perturbation kernels satisfy:
\begin{equation}
    q_{0t}(\vx_t | \vx_0) = \mathcal{N}(\vx_t | \alpha_t \vx_0, \sigma_t^2 \bm{I}),
\end{equation}
where $\alpha_t$ and $\sigma_t$ specify the noise schedule. While the transition kernels for VP SDE have the form:
\begin{equation}
    q(\vx_t|\vx_{t-1}) = \mathcal{N} \left( \frac{\alpha_{t}}{\alpha_{t-1}} \vx_{t-1}, \sqrt{1- \frac{\alpha_{t}^2}{\alpha_{t-1}^2}} \bm{I} \right).
\end{equation}
The detailed implementation of multistep consistency sampling in~\cite{song2023consistency} and our proposed strategic stochastic sampling are presented in~\cref{alg:mcs} and~\cref{alg:sss}, respectively. Both of them alternate between a denoising sub-step and a diffusing sub-step. The key distinction lies in the fact that the ending point of multistep consistency sampling is always set to 0. In contrast, strategic stochastic sampling introduces control through the parameter $\gamma$, allowing for the adjustment of the random noise level during alternating denoise and diffuse steps.

\begin{minipage}[t]{0.5\textwidth}
\begin{algorithm}[H]
   \caption{Multistep Consistency Sampling}
   \label{alg:mcs}
   \setstretch{1.02}
\begin{algorithmic}
   \STATE {\bfseries Input:} Consistency model $\vf_{\bm{\theta^*}} (\cdot, t)$, sequence of time points $T = \tau_1 > \tau_2 > \cdots > \tau_N$, initial value $\tilde{\vx}_{\tau_1} = \tilde{\vx}_T \sim \mathcal{N}(0, \bm{I})$ \\
   \FOR{$n=1$ {\bfseries to} $N-1$}
   \STATE Sample $\vz \sim \mathcal{N}(0, \bm{I})$
   \STATE Fixed ending time $\tau_n^{'} = 0$ \\
   {\color{blue}{\# Denoise}} \\
   \STATE $\vx_{\tau_n \rightarrow \tau_n^{'}} \leftarrow \vf_{\bm{\theta^*}}(\tilde{\vx}_{\tau_n}, \tau_n)$ \\
   {\color{blue}{\# Diffuse}} \\
   \STATE $\hat{\vx}_{\tau_{(n+1)}} \leftarrow \alpha_{\tau_{(n+1)}} \vx_{\tau_n \rightarrow \tau_n^{'}}  + \sigma_{\tau_{(n+1)}} \vz$
   \STATE
   \ENDFOR
   \STATE $\vx_{\tau_N \rightarrow 0} \leftarrow \vf_{\bm{\theta^*}}(\tilde{\vx}_{\tau_N}, \tau_N)$
   \STATE {\bfseries Output:} $\vx_{\tau_N \rightarrow 0}$
\end{algorithmic}
\end{algorithm}
\end{minipage} 
\begin{minipage}[t]{.5\linewidth}
\begin{algorithm}[H]
   \caption{Strategic Stochastic Sampling}
   \label{alg:sss}
\begin{algorithmic}
   \STATE {\bfseries Input:} Trajectory consistency model $\vf_{\bm{\theta^*}} (\cdot, t, s)$, sequence of time points $T = \tau_1 > \tau_2 > \cdots > \tau_N$, initial value $\tilde{\vx}_{\tau_1} = \tilde{\vx}_T \sim \mathcal{N}(0, \bm{I})$, parameter $\gamma \in [0,1]$ \\
   \FOR{$n=1$ {\bfseries to} $N-1$}
   \STATE Sample $\vz \sim \mathcal{N}(0, \bm{I})$
   \STATE Calculate ending time $\tau_n^{'} = (1- \gamma) \tau_{(n+1)}$ \\
   {\color{blue}{\# Denoise}} \\
   \STATE $\vx_{\tau_n \rightarrow \tau_n^{'}} \leftarrow \vf_{\bm{\theta^*}}^{\rightarrow  \tau_n^{'}}(\hat{\vx}_{\tau_n}, \tau_n)$ \\
   {\color{blue}{\# Diffuse}} \\
   \STATE $\tilde{\vx}_{\tau_{(n+1)}} \leftarrow \frac{\alpha_{\tau_{(n+1)}}}{\alpha_{\tau_n^{'}}} \vx_{\tau_n \rightarrow \tau_n^{'}}  + \sqrt{1- \frac{\alpha_{\tau_{(n+1)}}^2}{\alpha_{\tau_n^{'}}^2}} \vz$
   \ENDFOR 
   \STATE $\vx_{\tau_N \rightarrow 0} \leftarrow \vf_{\bm{\theta^*}}^{\rightarrow 0}(\tilde{\vx}_{\tau_N}, \tau_N)$
   \STATE {\bfseries Output:} $\vx_{\tau_N \rightarrow 0}$
\end{algorithmic}
\end{algorithm}
\end{minipage}

\section{Implementation Details} \label{app:imp_details}

\subsection{Dataset}
We trained TCD on the LAION5B High-Res dataset~\cite{schuhmann2022laion}. Images with an aesthetic score~\cite{ch2022clip+mlp} lower than 5.8 were filtered out. For training, all images were initially resized to 1024 pixels by the shorter side and subsequently randomly cropped to dimensions of $1024 \times 1024$.

\subsection{Hyper-parameters}
In our experiments, we utilized the AdamW optimizer with $\beta_1 = 0.9, \beta_2 = 0.999$, and a weight decay of 0.01. For TCF(2) parameterization, the itermidiate timestep $u$ in~\cref{eq:tcd-2} was set as $u := t_\lambda (\lambda_s + \lambda_t)$, where $t_\lambda$ is the inverse function of $\lambda_t$. For TCF(S+) parameterization, to encode the ending time $s$ into the TCD model, we create sinusoidal timestep embeddings matching the implementation in~\cite{ho2020denoising}. Subsequently, we project it with a simple MLP, integrating it into the TCD backbone by adding the projected $s$-embedding to the start time $t$-embedding, following the approach in the original Stable Diffusion models~\cite{rombach2022high}. We choose the DDIM~\cite{song2020denoising} as the ODE solver and set the skipping step $k = 20$ in~\cref{eq:k_solver}. During training, we initialize the trajectory consistency function with the same parameters as the teacher diffusion model. The guidance scale for guided distillation~\cite{meng2023distillation} was set to $[2, 14]$. We did not utilize the EDM's skip scale and output scale, as we found it did not provide any benefit. For TCF(1) and TCF(2), the batch size was set to 256, and the learning rate was set to 4.5e-6, and the LoRA rank was 64. We did not use EMA for TCF(1) and TCF(2). For TCF(S+), the batch size was set to 192, the learning rate was set to 4.0e-6. We did not use LoRA for TCF(S+) but directly fine-tuned the U-Net, and the EMA decay rate was set to 0.95.

\subsection{Hardware and Efficiency}
All our models were trained using 8 80G A800 units. TCF(1) was trained for 3000 iterations, taking 15 hours, demonstrating high training efficiency. TCF(2) employed the same parameterization method as TCF(1) during training. TCF(S+) was trained for 43,000 iterations, spanning 5 days and 20 hours. The low convergence rate of TCF(S+) may be attributed to the introduction of additional parameters, resulting in low distillation efficiency of teacher information. We defer the exploration of improved design for enhancing the efficiency of TCF(S+) to future work. The inference time in~\cref{fig:teaser} was tested on an A800, averaged over 16 prompts.

\section{Additional Results}
\subsection{More Samples with Different NFEs} \label{app:diff_nfes}
\begin{figure*}[ht!]
    \centering
    \includegraphics[width=0.68\linewidth]{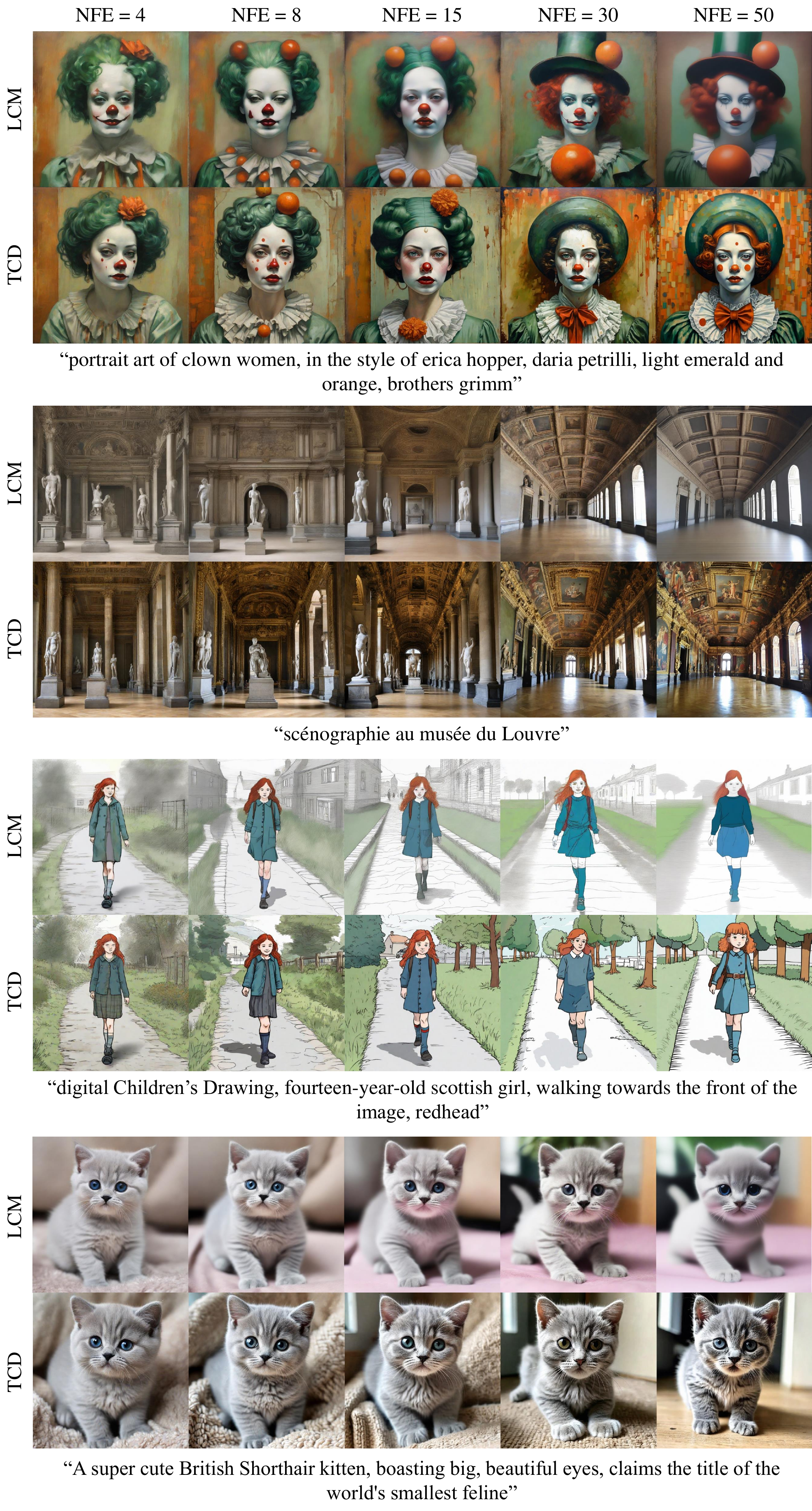}
    \caption{More samples with different NFEs.}
    \label{fig:diff_nfes_app}
\end{figure*}

In~\cref{fig:diff_nfes_app}, we present additional samples synthesized by LCM and TCD with varying NFEs. These samples consistently demonstrate that the details of images generated by LCM tend to vanish as the NFEs increase, whereas TCD either maintains or enhances image details at higher NFEs.

\subsection{More Samples with Different Stochastic Parameter $\gamma$} \label{app:diff_gamma}
\begin{figure*}[ht!]
    \centering
    \includegraphics[width=\linewidth]{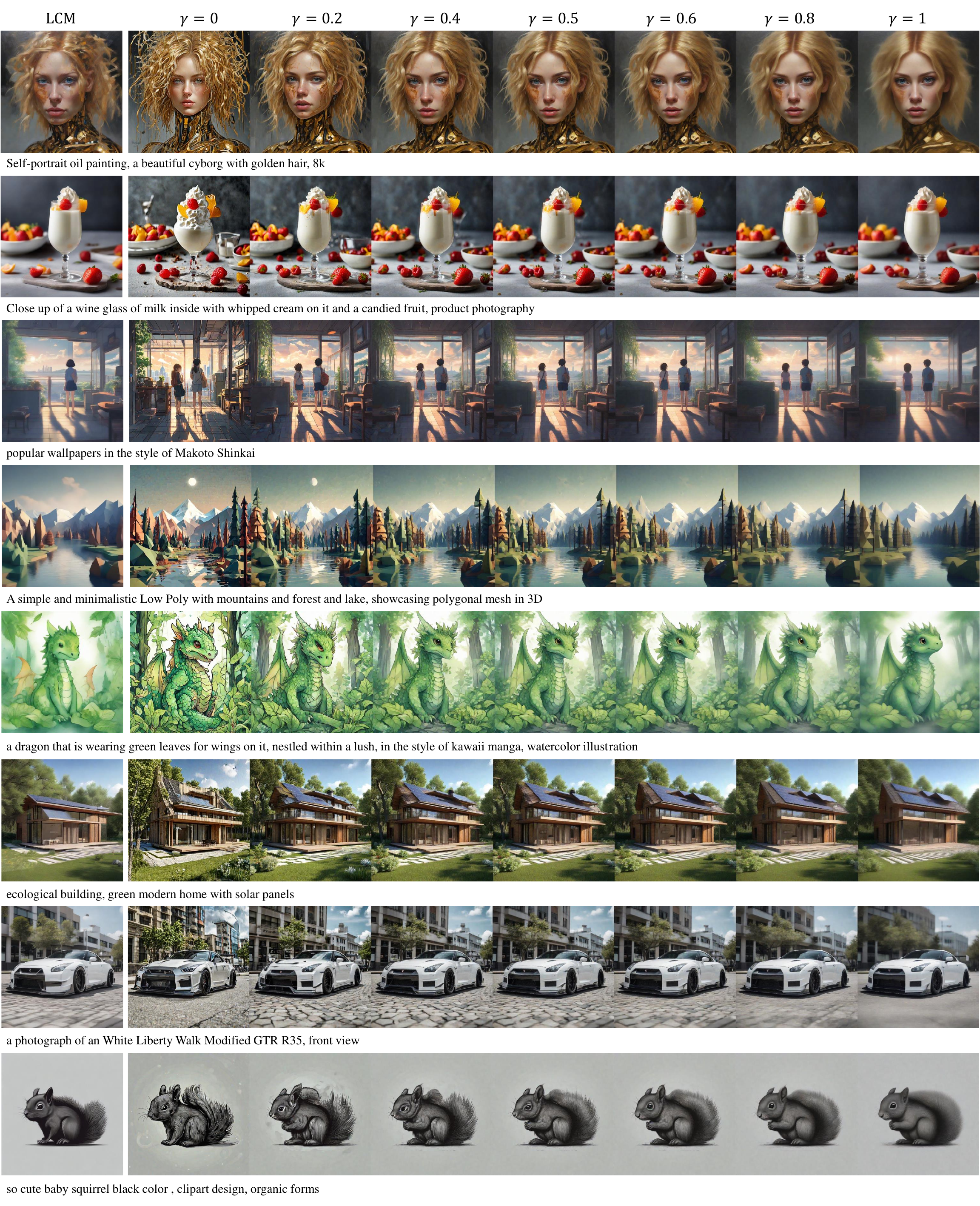}
    \caption{More samples with different stochastic parameter $\gamma$.}
    \label{fig:diff_gamma_app}
\end{figure*}

In~\cref{fig:diff_gamma_app}, we showcase additional samples generated by TCD with varying stochastic parameter $\gamma$, demonstrating that the visual complexity and quality of the image gradually improve as $\gamma$ increases.

\subsection{More Comparisons Results} \label{app:more_compare}
\begin{figure*}[ht!]
    \centering
    \includegraphics[width=\linewidth]{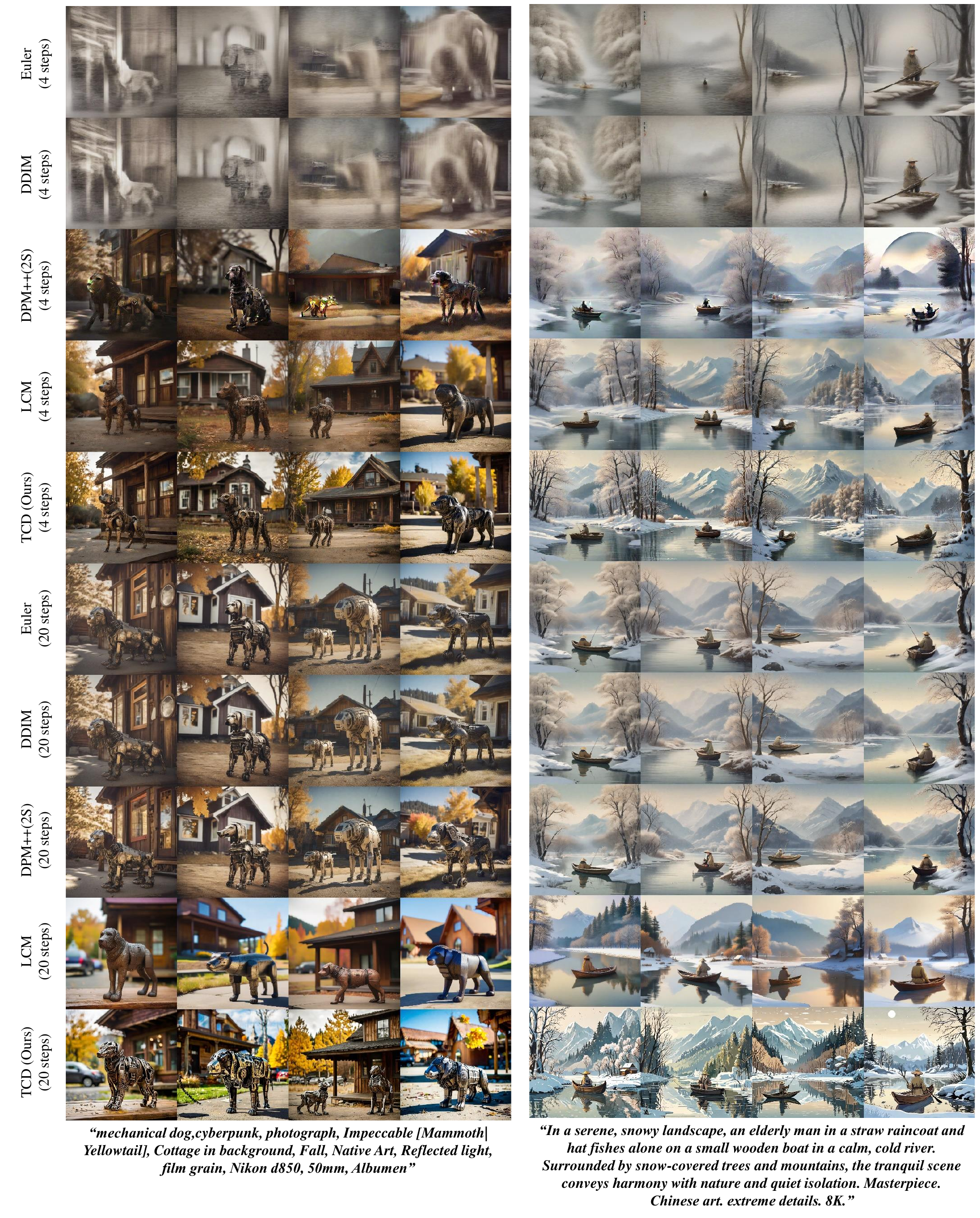}
    \caption{More comparisons results.}
    \label{fig:cr1}
\end{figure*}

\begin{figure*}[ht!]
    \centering
    \includegraphics[width=\linewidth]{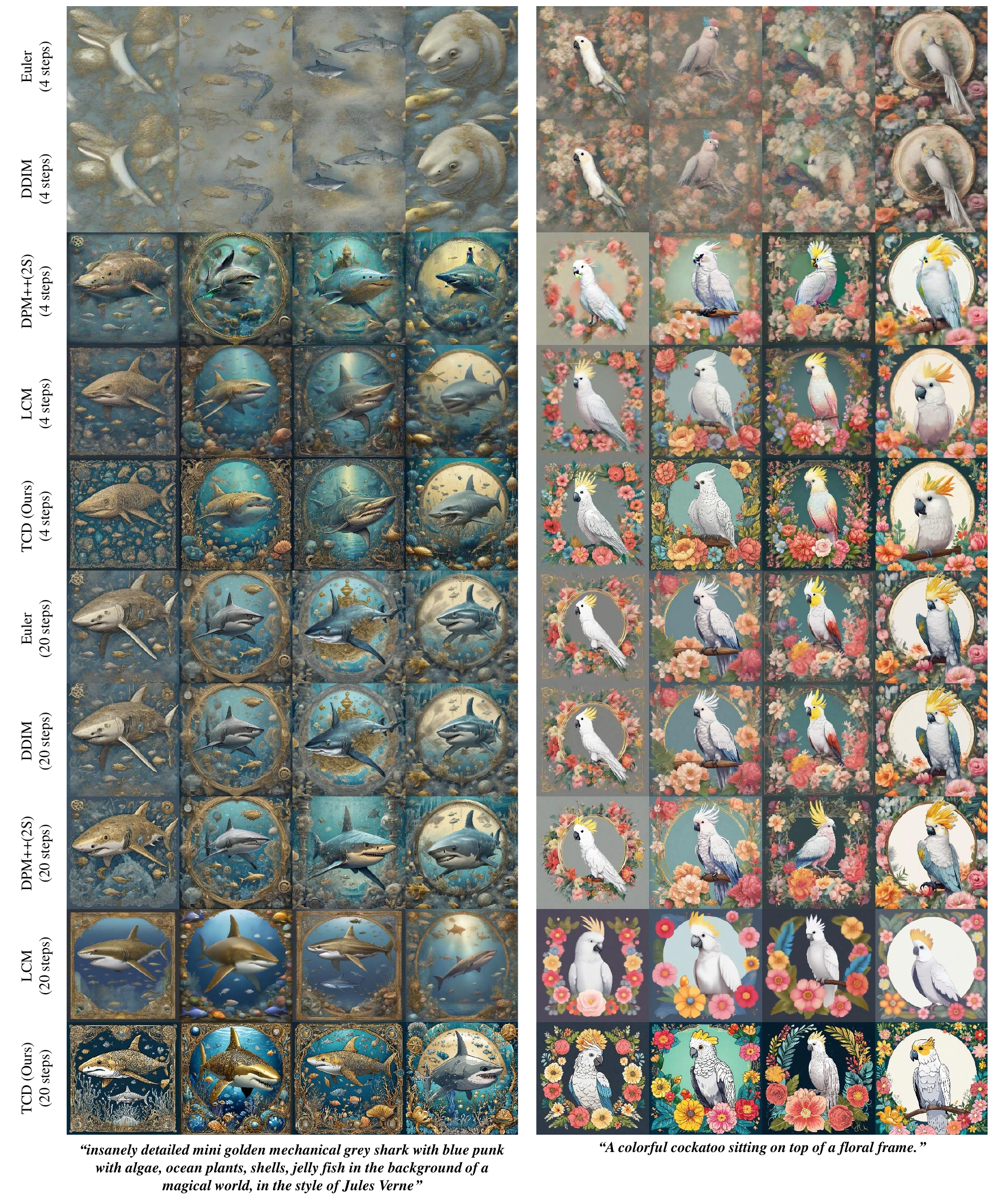}
    \caption{More comparisons results.}
    \label{fig:cr2}
\end{figure*}

\begin{figure*}[ht!]
    \centering
    \includegraphics[width=\linewidth]{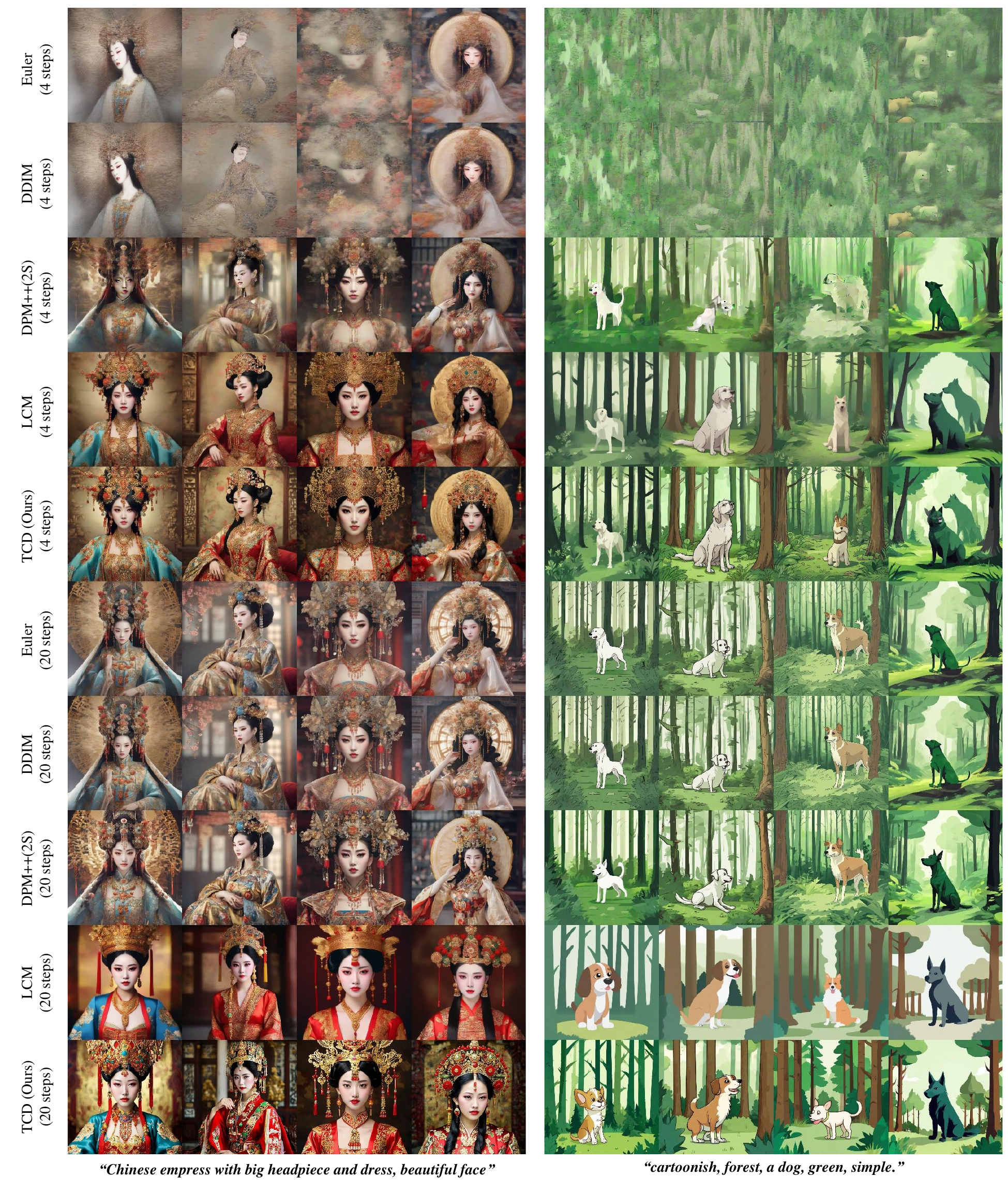}
    \caption{More comparisons results.}
    \label{fig:cr3}
\end{figure*}

\begin{figure*}[ht!]
    \centering
    \includegraphics[width=\linewidth]{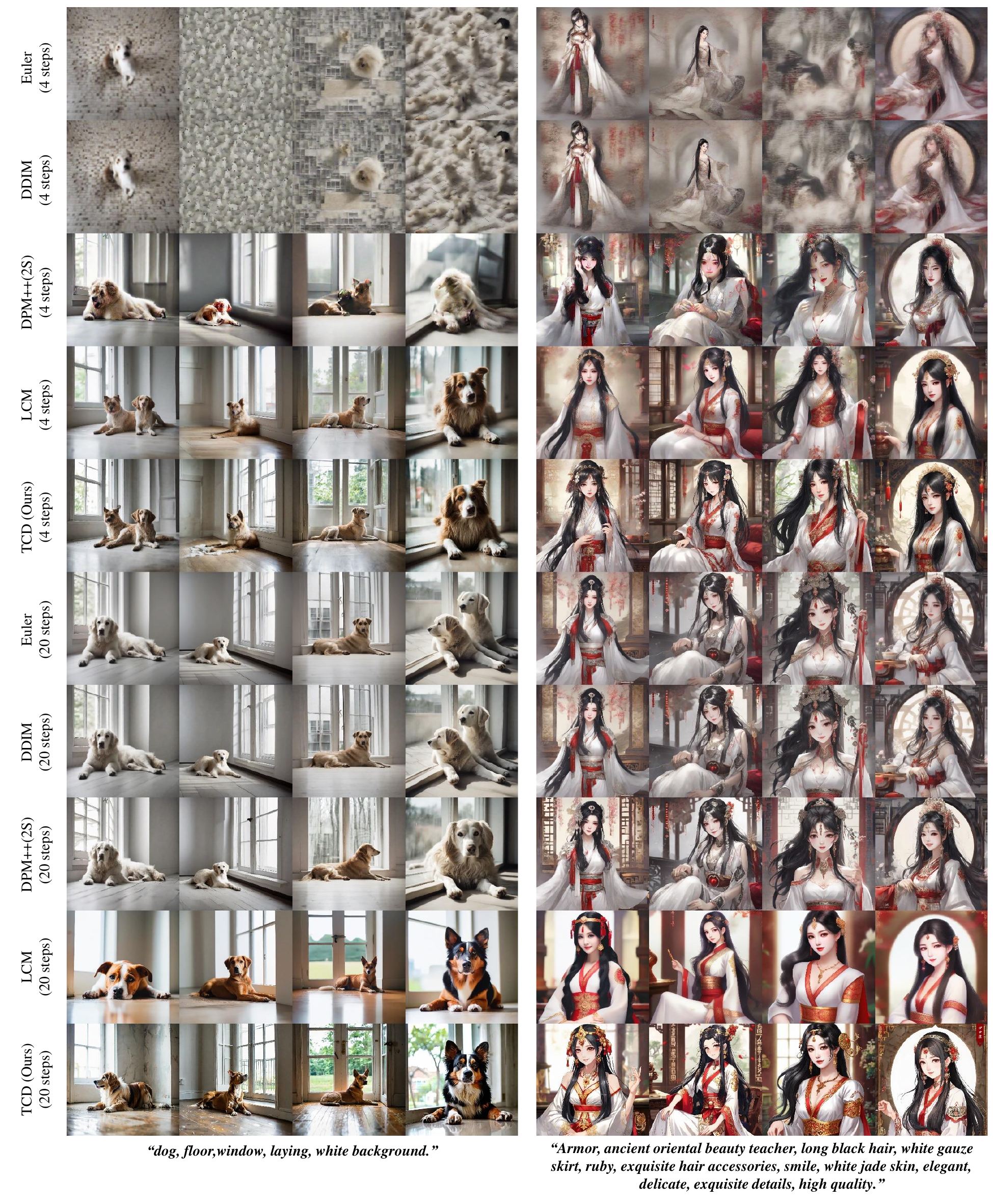}
    \caption{More comparisons results.}
    \label{fig:cr4}
\end{figure*}

\begin{figure*}[ht!]
    \centering
    \includegraphics[width=\linewidth]{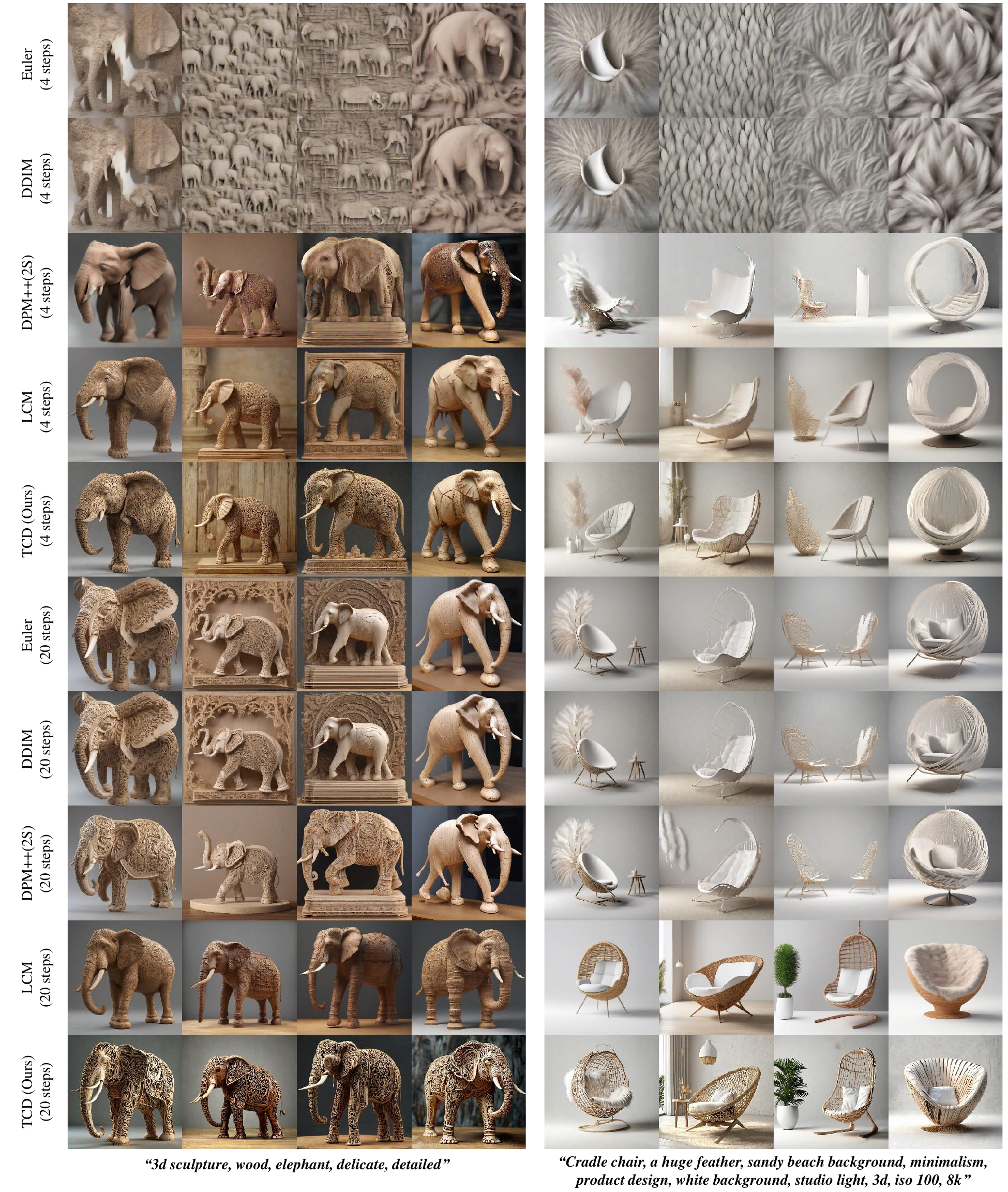}
    \caption{More comparisons results.}
    \label{fig:cr5}
\end{figure*}

In~\cref{fig:cr1} to~\cref{fig:cr5}, we present additional examples for comparison with state-of-the-art methods.
These samples illustrate that our TCD can generate high-quality images with few NFEs, outperforming LCM. Additionally, it can produce more detailed and high-definition images with sufficient NFEs, even surpassing the performance of DPM-Solver++(2S).

\subsection{More Samples from TCD's Versatility Testing} \label{app:versatility_samples}
\begin{figure*}[ht!]
    \centering
    \includegraphics[width=0.72\linewidth]{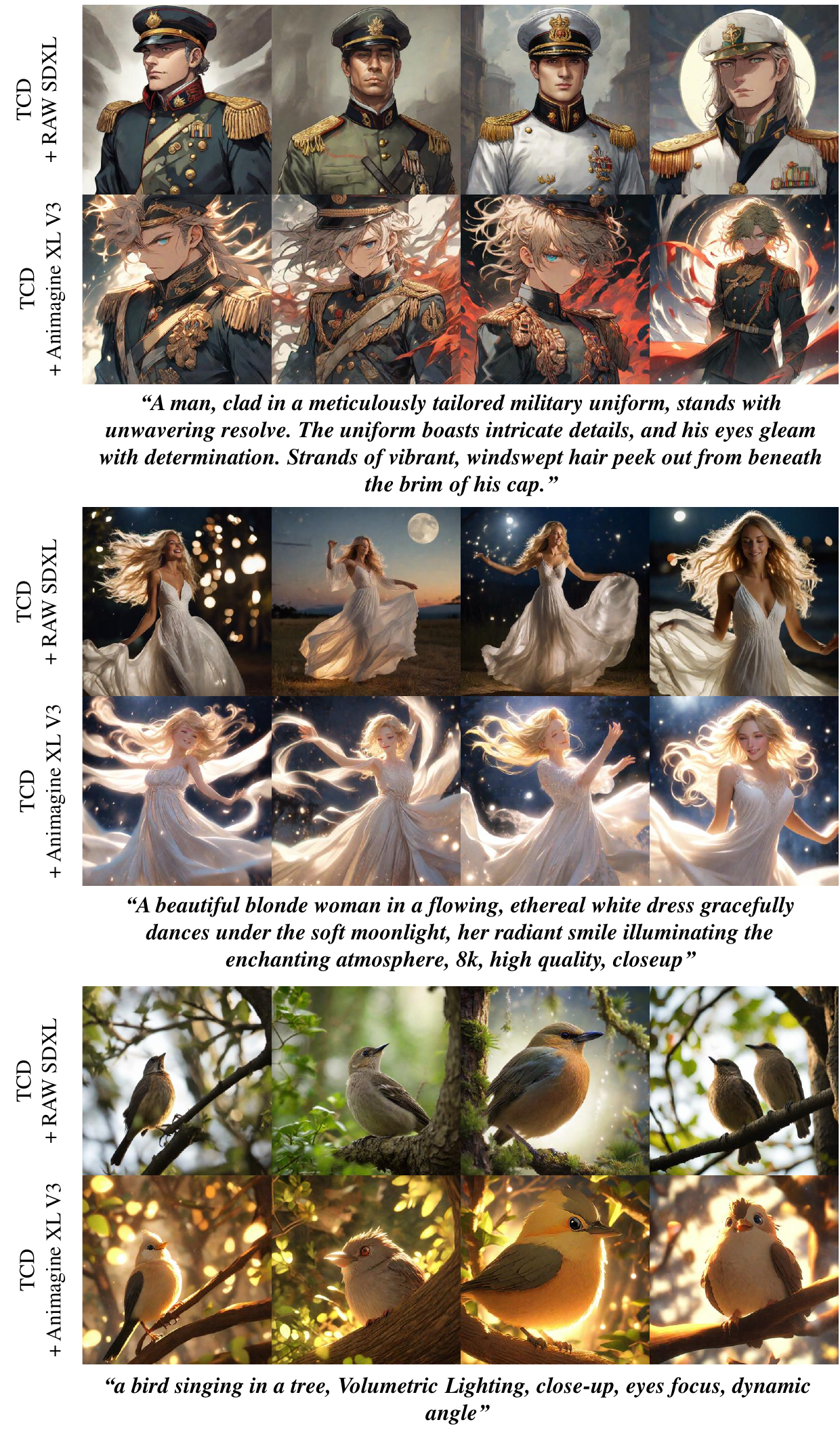}
    \caption{Qualitative results of TCD using different base models: SDXL and Animagine XL V3. It is worth noting that we employed \textit{the same TCD parameters} for both models. All samples were generated using \textit{4 steps}. In each subfigure, the top row corresponds to TCD + SDXL, and the bottom row corresponds to TCD + Animagine XL V3.}
    \label{fig:animagine}
\end{figure*}

\begin{figure*}[ht!]
    \centering
    \includegraphics[width=0.72\linewidth]{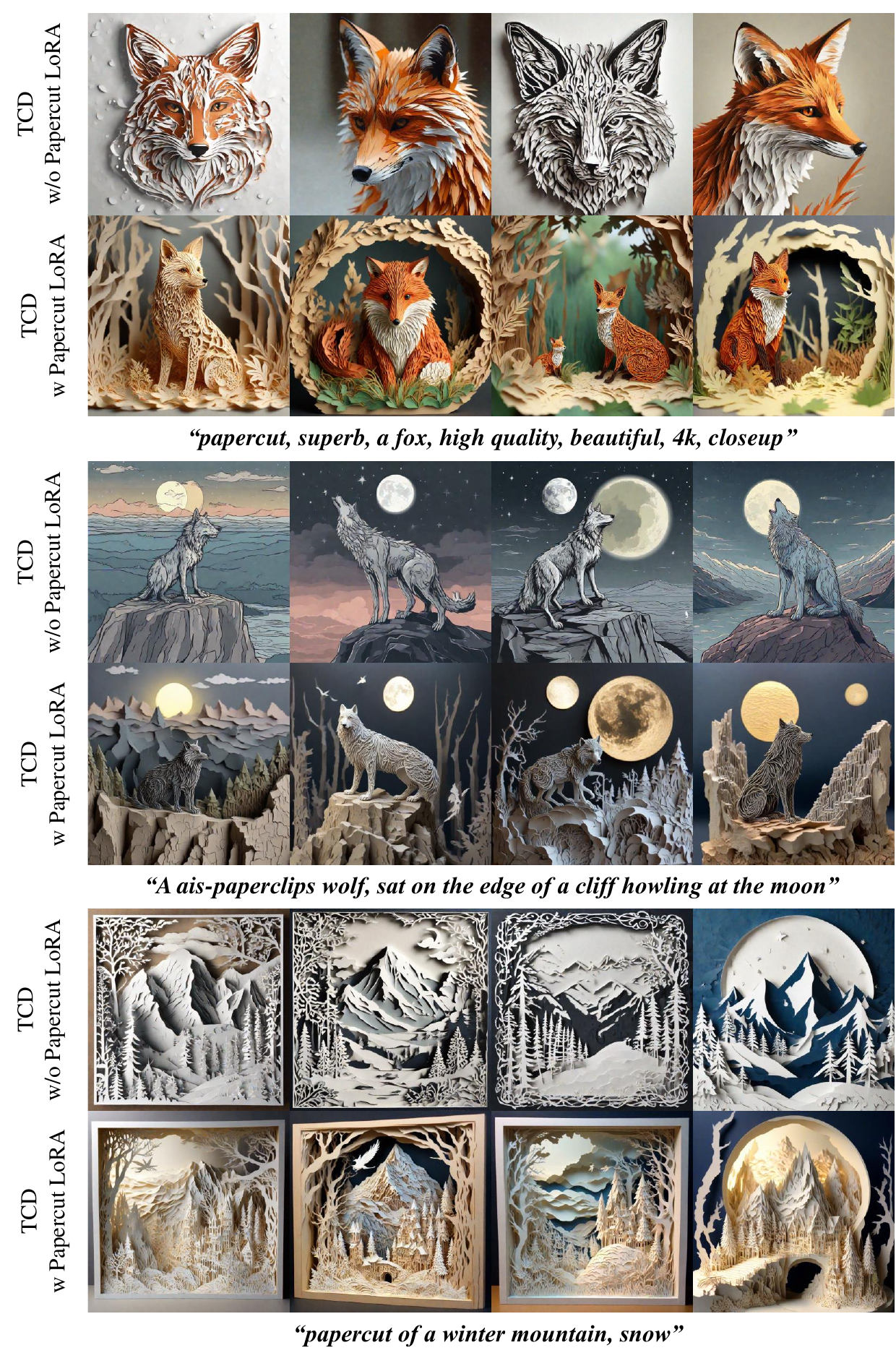}
    \caption{Qualitative results of TCD with and without Papercut XL LoRA. We used \textit{the same TCD parameters}. All samples are generated using \textit{4 steps}. In each subfigure, the top row corresponds to TCD without Papercut LoRA, and the bottom row corresponds to TCD with Papercut LoRA. The Lora scale of Papercut was set to 1.0 in the experiments.}
    \label{fig:papercut}
\end{figure*}

\begin{figure*}[ht!]
    \centering
    \includegraphics[width=0.72\linewidth]{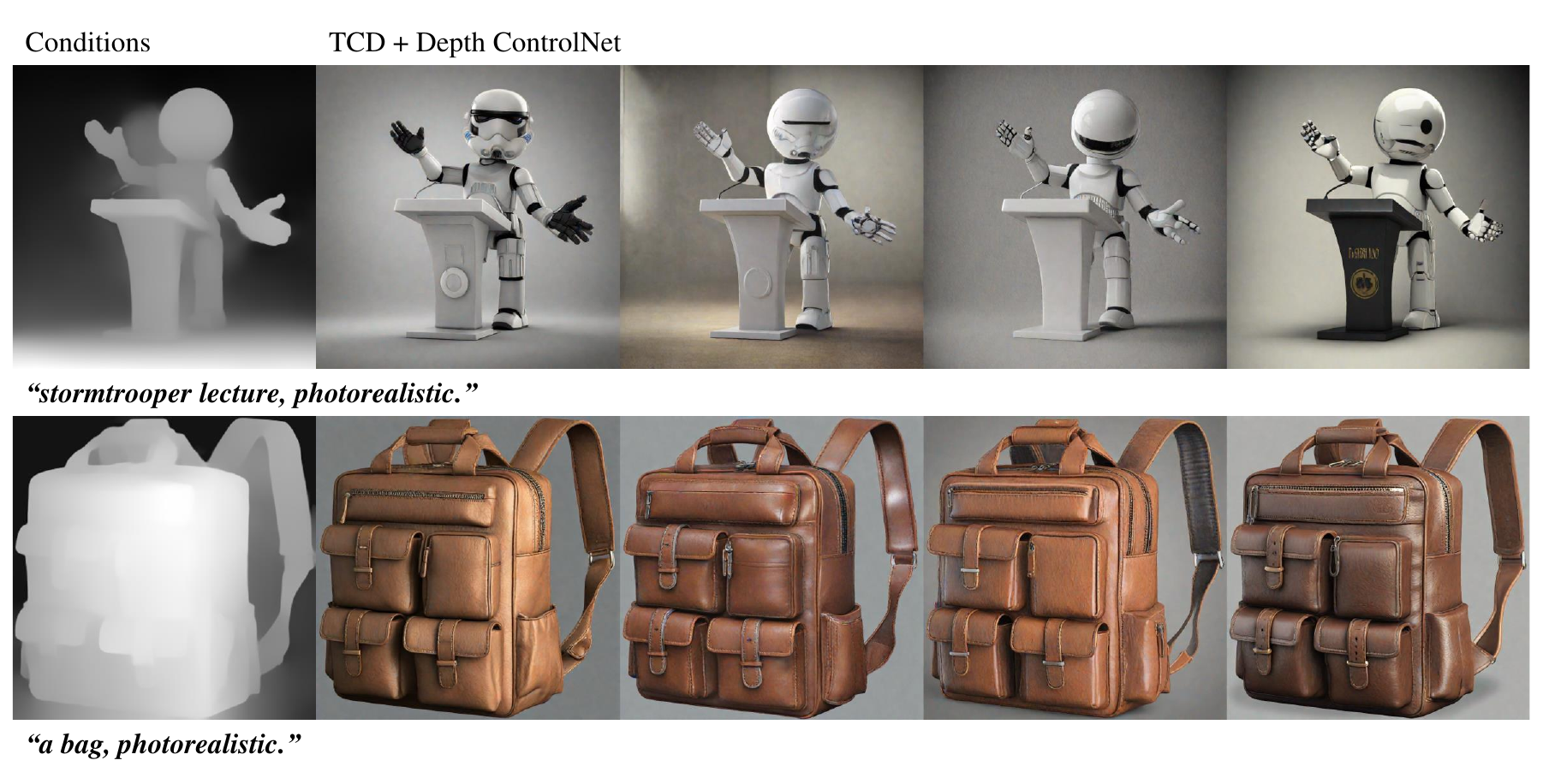}
    \caption{Qualitative results of TCD with Depth ControlNet. All samples are generated using \textit{4 steps}.}
    \label{fig:controlnet_depth}
\end{figure*}

\begin{figure*}[ht!]
    \centering
    \includegraphics[width=0.72\linewidth]{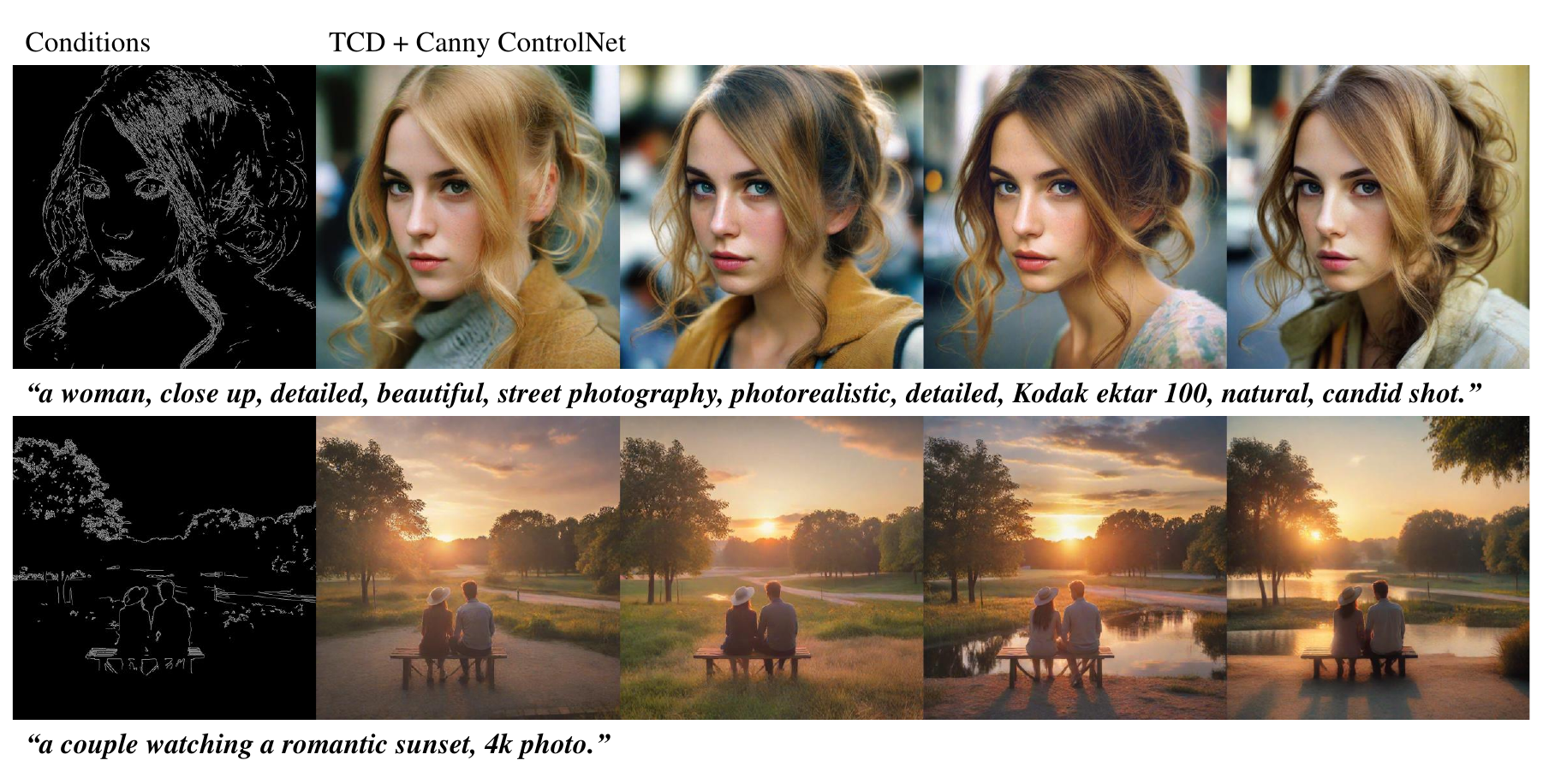}
    \caption{Qualitative results of TCD with Canny ControlNet. All samples are generated using \textit{4 steps}.}
    \label{fig:controlnet_canny}
\end{figure*}

\begin{figure*}[ht!]
    \centering
    \includegraphics[width=0.72\linewidth]{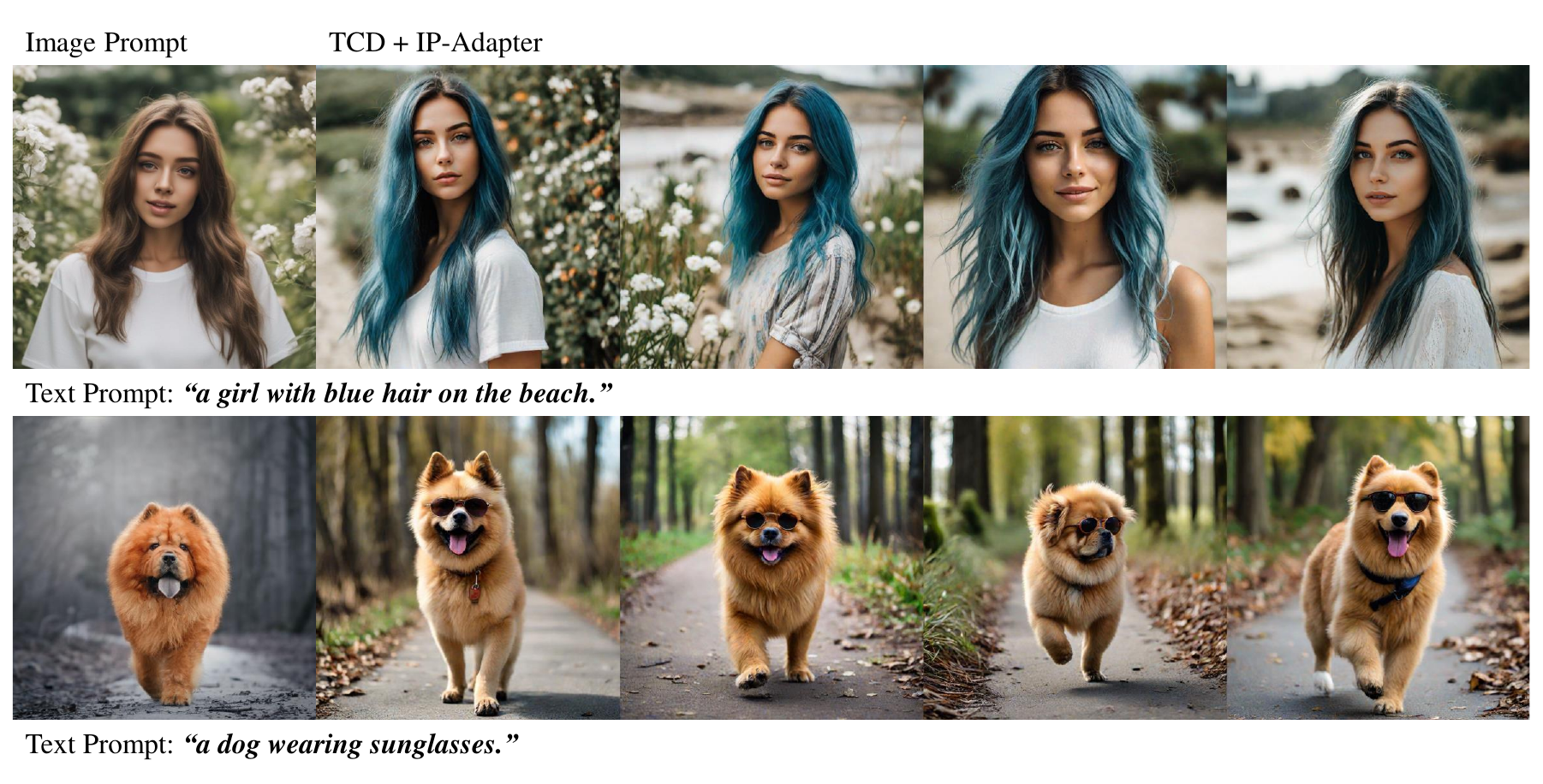}
    \caption{Qualitative results of TCD with IP-Adapter. All samples are generated using \textit{4 steps}.}
    \label{fig:ip_adapter}
\end{figure*}

To ascertain the versatility of TCD, we test it on a wide range of models, including the popular community model Animagine XL V3~\footnote{The checkpoint of \textit{Animagine} can be found in \texttt{https://civitai.com/models/260267/animagine-xl-v3}}, LoRA Papercut XL~\footnote{The checkpoint of \textit{Papercut} can be found in \texttt{https://civitai.com/models/122567/papercut-sdxl}}, Depth ControlNet~\footnote{\textit{Depth ControlNet}: \texttt{https://huggingface.co/diffusers/controlnet-depth-sdxl-1.0}}, Canny ControlNet~\footnote{\textit{Canny ControlNet}: \texttt{https://huggingface.co/diffusers/controlnet-canny-sdxl-1.0}}, and IP-Adapter~\footnote{\textit{IP-Adapter}: \texttt{https://github.com/tencent-ailab/IP-Adapter}}.
In~\cref{fig:animagine} to~\cref{fig:ip_adapter}, we observe that TCD can be directly applied to all these models, accelerating their sampling with only \textit{4 steps}. It is worth noting that, in this experiment, all models share \textit{the same TCD LoRA parameters}.



\end{document}